\documentclass{IEEEojcsys}

\usepackage[colorlinks,urlcolor=blue,linkcolor=blue,citecolor=blue]{hyperref}

\usepackage{color,array}

\usepackage{graphicx}

\usepackage{textcomp}
\usepackage{stfloats}
\usepackage{url}
\usepackage{verbatim}
\usepackage{cite}
\usepackage{dsfont}
\usepackage{bm}  
\usepackage{mathtools}
\usepackage{caption}
\usepackage{subcaption}
\usepackage{floatrow}
\usepackage{xcolor}
\usepackage{amsmath,amsfonts,amsfonts, amssymb, graphicx, amsthm}
\usepackage{algorithm}
\usepackage{algpseudocode}
\usepackage{array}
\usepackage{booktabs}
\usepackage{multirow}
\usepackage{rotating}
\usepackage{makecell}
\usepackage{hyperref}

\newcommand{\bxi}{\bm{\xi}}
\newcommand{\bpsi}{\bm{\psi}}
\newcommand{\bmu}{\bm{\mu}}
\newcommand{\bz}{\bm{z}}
\newcommand{\bmeta}{\bm{\eta}}
\newcommand{\bwmeta}{\widetilde{\bm{\eta}}}

\newcommand{\e}{\mathbb{E}}
\newcommand{\f}{\mathcal{F}}
\newcommand{\bmf}{\boldsymbol{\mathcal{F}}}
\newcommand{\w}{\bm{w}}
\newcommand{\m}{\bm{m}}
\newcommand{\bde}{\bm{\delta}}
\newcommand{\rb}{\bm{r}}
\newcommand{\s}{\bm{s}}
\newcommand{\gb}{\bm{g}}
\newcommand{\ac}{\bm{a}}
\newcommand{\bnu}{\bm{\nu}}

\newcommand{\bH}{\bm{H}}
\newcommand{\bD}{\bm{d}}
\newcommand{\bvart}{\bm{\vartheta}}
\newcommand{\dkl}{D_{\textup{KL}}}

\newcommand{\btv}{B_{\textup{TV}}}
\newcommand{\wbtv}{\widetilde{B}_{\textup{TV}}}
\newcommand{\mcl}{\text{col}}
\newcommand{\mdiag}{\text{diag}}
\newcommand{\bT}{\mathbb{T}}
\newcommand{\rma}{R_{\textup{max}}}
\newcommand{\bfi}{B_{\phi}}
\newcommand{\lfi}{L_{\phi}}

\newcommand{\bmch}{\bm{\mathcal{H}}} 
\renewcommand{\qed}{\hfill\blacksquare}

\newenvironment{myproof}
{\noindent \textit{Proof.}}{$\qed$ \\}

\DeclareMathOperator{\T}{\mathsf{T}}

\DeclareMathOperator{\cw}{{\scriptstyle\mathcal{W}}}
\DeclareMathOperator{\bcw}{{\boldsymbol{\scriptstyle\mathcal{W}}}}

\newtheorem{assumption}{Assumption}
\newtheorem{theorem}{Theorem}
\newtheorem{corollary}{Corollary}
\newtheorem{lemma}{Lemma}

 \algdef{SE}[FORA]{FORA}{ENDFORA}[1]{\algorithmicfor\ #1\ \textbf{locally}}{\algorithmicend\ \algorithmicfor}
  \algdef{SE}[FORB]{FORB}{ENDFORB}[1]{\algorithmicfor\ #1\ \textbf{evolve}}{\algorithmicend\ \algorithmicfor}
   \algdef{SE}[FORC]{FORC}{ENDFORC}[1]{\algorithmicfor\ #1\ \textbf{adapt} \text{and} \textbf{combine}}{\algorithmicend\ \algorithmicfor}
    \algdef{SE}[FORD]{FORD}{ENDFORD}[1]{\algorithmicfor\ #1\ \textbf{combine} \text{and} \textbf{evolve} }{\algorithmicend\ \algorithmicfor}
     \algdef{SE}[FORE]{FORE}{ENDFORE}[1]{\algorithmicfor\ #1\ \textbf{combine}}{\algorithmicend\ \algorithmicfor}
          \algdef{SE}[FORF]{FORF}{ENDFORF}[1]{\algorithmicfor\ #1\ \textbf{update} \text{locally}}{\algorithmicend\ \algorithmicfor}

\begin{document}

\sptitle{Article Category}

\title{Policy Evaluation in Decentralized POMDPs with Belief Sharing} 


\author{MERT KAYAALP\affilmark{1}}

\author{FATIMA GHADIEH\affilmark{2}}

\author{ALI H. SAYED\affilmark{1}}

\affil{Adaptive Systems Laboratory, \'{E}cole Polytechnique F\'{e}d\'{e}rale de Lausanne (EPFL), CH-1015, Lausanne, Switzerland} 
\affil{American University of Beirut, Beirut 1107 2020, Lebanon} 

\corresp{CORRESPONDING AUTHOR: M. Kayaalp. E-mails: \href{mailto:mert.kayaalp@epfl.ch}{mert.kayaalp@epfl.ch}}
\authornote{The work of F. Ghadieh was performed while she was a student intern at the Adaptive Systems Laboratory, EPFL.}

\markboth{POLICY EVALUATION IN DECENTRALIZED POMDPs WITH BELIEF SHARING}{KAYAALP {\itshape ET AL}.}

\begin{abstract}
Most works on multi-agent reinforcement learning focus on scenarios where the state of the environment is fully observable. In this work, we consider a cooperative policy evaluation task in which agents are not assumed to observe the environment state directly. Instead, agents can only have access to noisy observations and to belief vectors. It is well-known that finding global posterior distributions under multi-agent settings is generally NP-hard. As a remedy, we propose a fully decentralized belief forming strategy that relies on individual updates and on localized interactions over a communication network. In addition to the exchange of the beliefs, agents exploit the communication network by exchanging value function parameter estimates as well. We analytically show that the proposed strategy allows information to diffuse over the network, which in turn allows the agents' parameters to have a bounded difference with a centralized baseline. A multi-sensor target tracking application is considered in the simulations.
\end{abstract}


\maketitle

\section{INTRODUCTION}

Multi-agent reinforcement learning (MARL) \cite{busoniu2008comprehensive,zhang2021multi} is a useful paradigm for determining optimal policies in sequential decision making tasks involving a group of agents. MARL has been applied successfully in several contexts, including sensor networks \cite{huang2010distributed,lunden2013multiagent}, team robotics \cite{bhattacharya2021multiagent}, and video games \cite{vinyals2019grandmaster,samvelyan2019}. MARL owes this success in part to recent developments in better function approximators such as deep neural networks \cite{lecun2015deep}. 

Many works on MARL focus on the case where agents can directly observe the global state of the environment. However, in many scenarios, agents can only receive partial information about the state. The decentralized partially observable Markov decision process (Dec-POMDP) framework \cite{oliehoek2016concise} is applicable to these types of situations. However, a large body of MARL work assumes that Dec-POMDPs observe data that are deterministic and known functions of the underlying state, which is not the case in general. Consider, for example, robots that receive noisy observations from their sensors. The underlying observation model is stochastic in this case.

Under stochastic observation models, one common strategy is to keep track of the posterior distribution (belief) over the set of states, which is known to be a sufficient statistic of the history of the system \cite{sondik1978optimal,kaelbling1998}. For single agents, this posterior distribution can be obtained at each iteration with the optimal Bayesian filtering recursion \cite{krishnamurthy_2016}. Unfortunately, for multi-agent systems, forming this global posterior belief requires aggregation of all data from across all agents in general. The agents can form it in a distributed manner only when they have access to the private information from other agents in the network. And even when agents have access to this level of global knowledge, the computational complexity of forming the global posterior distribution is known to be NP-hard \cite{hazla2021bayesian} in addition to its large memory requirements. Moreover, obtaining beliefs necessitates significant knowledge about the underlying model of the environment, which is generally not available in practice.

Therefore, instead of forming beliefs, most MARL algorithms \cite{omidshafiei2017deep,gupta2017cooperative,foerster2018counterfactual} resort to a model-free and end-to-end approach where agents try to simultaneously learn a policy and an embedding of the history that can replace the beliefs (e.g., recurrent neural networks (RNNs)). Nevertheless, recent empirical works suggest that this model-free approach can be sub-optimal when the underlying signals of the environment are too weak to train a model such as RNN \cite{moreno2018neural,gregor2018temporal}. Moreover, RNNs (or alternative machine learning models) are usually treated as black boxes. In other words, these algorithms lack model interpretability, which is critical for trustworthy systems (see \cite{sayed_2022}). Furthermore, even though end-to-end approaches have shown remarkable performance empirically, they are still based on heuristics and lack theoretical guarantees on their performance. Compared to modular approaches, they are inefficient in terms of adaptability and generalization to similar tasks.

As an alternative, there is a recent interest towards improving belief-based MARL approaches \cite{moreno2021neural,muglich2022,mao2020}. These works have focused on emulating conventional beliefs with generative models, or with models learned from action/observation trajectories (in a supervised fashion). In this paper, we also examine belief-based strategies for MARL. In particular, we are interested in the multi-agent policy evaluation problem. Our work complements \cite{moreno2021neural,muglich2022,mao2020} in the sense that we assume that agents are already capable of forming \emph{local} beliefs with sufficient knowledge (i.e., with learned \emph{local} likelihood and transition models) or with generative models. Our focus is on the challenge of approximating the \emph{global} Bayesian posterior in a \emph{distributed} manner.

\noindent \textbf{Contributions}.
\begin{itemize}
    \item We consider a setting where agents only get partial observations from the underlying state of nature, as opposed to prior work on MARL over networks \cite{kar2013,zhang2018fully,cassano2021,macua2021fully,sha2022fully,lin_stochastic_network_2021,wang2020,sun2020finite,lin2021} that assume agents have full state information. Moreover, as opposed to the literature on decentralized stochastic control \cite{mahajan2016decentralized,malikopoulos2022team,yuksel2009stochastic,nayyar2013decentralized}, in our setting, agents need to learn their value functions from data. More specifically, in our Dec-POMDP framework, agents only know their local observations, actions, and rewards but they are allowed to communicate with their immediate neighbors over a graph. In the proposed strategy (Algorithm~\ref{alg:fully_decent}), agents exchange both their belief and value function estimates. 
    \item We show in Theorem~\ref{th:kl_without_net_assumption} that by exchanging beliefs, agents keep a bounded disagreement with the global posterior distribution, which requires fusing all observations and actions. Also, exchanging value function parameters enables agents
     to cluster around the network centroid for sufficiently small learning rates (Theorem~\ref{th:network_disagreement}). Furthermore, we prove that the network centroid attains a bounded difference with a strategy that requires centralized training (Theorem~\ref{th:centralized_disagreement}). 
    \item By means of simulations, we illustrate that agents attain a small mean-square distance from the network centroid. Moreover, the squared Bellman error (SBE) averaged over the network is shown to be comparable to the SBE of the centralized strategy.
\end{itemize}

\noindent \textbf{Paper Organization}. In Sec.~\ref{sec:related_work}, we present additional related work. In Sec.~\ref{sec:preliminaries}, for ease of exposition and introducing notation, we describe the problem in single-agent setting. In Sec.~\ref{sec:multi_agent}, we propose algorithms for multi-agent policy evaluation. Sec.~\ref{sec:theoretical_results} includes the theoretical results, and Sec.~\ref{sec:simulation_results} includes numerical simulations. 

\section{OTHER RELATED WORK}\label{sec:related_work}

 Our proposed strategy is based on temporal-difference (TD) learning \cite{sutton2018reinforcement,tsitsiklis1996analysis}, and makes use of function approximation. TD-learning for POMDPs are considered in \cite{singh1994learning,rodriguez1999}, and function approximations are incorporated in \cite{kimura1997reinforcement,cai2022reinforcement}, albeit in single-agent setting. The main contribution of the present work is to the networked multi-agent setting.
    
A plethora of work studies decentralized policy evaluation over networks \cite{kar2013,zhang2018fully,cassano2021,macua2021fully,sha2022fully,lin_stochastic_network_2021,wang2020,sun2020finite,lin2021}. Distributed versions of the TD-learning with linear function approximations are considered in \cite{wang2020,sun2020finite,lin2021}. However, these works assume that either the global state, or a deterministic function of it, is available to all agents. They overlook the stochastic nature of observations that takes place in many real-world applications. Also in deterministic setting, the works \cite{li2021,wang2021distributed} examine distributed linear quadratic control task when agents can observe local states only. In particular, \cite{wang2021distributed} proposes a cooperative strategy for tracking the global state that exploits networked communication between agents. However, in this strategy, global state estimation at each iteration is independent of the previous estimations. It ignores the correlation between consecutive states. Furthermore, communication between the agents is utilized only for global state estimation, and not utilized for local Q-function estimate sharing. In contrast, in the present work, \((i)\) observations are stochastic, \((ii)\) agents take advantage of the transition model of the state, and \((iii)\) they exchange value function parameters with their neighbors as well.

Our work is also related to the field of decentralized stochastic control \cite{mahajan2016decentralized} and dynamic team theory \cite{malikopoulos2022team}. This field studies problems in which different decision-makers have access to different sets of information while working towards a common team goal. Typically, these problems are defined by an information structure that specifies which agents have access to which pieces of information (e.g., observations or actions) \cite{mahajan2012information,saldi2022geometry}. Some approaches to solving these problems rely on the common information that arises from partial history sharing to \emph{all} other agents \cite{nayyar2013decentralized,arabneydi2015reinforcement,nayyar2019common}. In our networked setting, agents exchange value function parameters or beliefs at each iteration, without explicitly exchanging raw data, with their \emph{immediate} neighbors only. Nonetheless, repeated application of this procedure causes information to mix and diffuse throughout the whole network. Moreover, most existing works in the decentralized stochastic control literature assume full model knowledge of the system, whereas we consider the case of learning from data since the reward model is not known a priori. Also, sharing value function parameters and beliefs instead of raw data makes our algorithm advantageous in terms of privacy and scalability. A similar approach is considered in \cite{yuksel2009stochastic}, where the author proposes a belief-sharing pattern for decentralized control, rather than explicit information sharing as in prior work. However, they use a belief propagation algorithm over acyclic graphs, while we use a diffusion-based belief-sharing algorithm over cyclic networks. In addition, \cite{yuksel2009stochastic} considers the planning problem only whereas in this work we consider the policy evaluation problem, which requires learning from data.

For constructing local beliefs that approximate the global Bayesian posterior, we extend the diffusion HMM strategy (DHS) \cite{kayaalp2022dslw,kayaalp_dist_bayesian}. This algorithm requires only one round of communication per state change, as opposed to other strategies \cite{hlinka_2012,battistelli_2014} that require multiple rounds of communication until network consensus at each iteration. Also, in contrast to other distributed Bayesian filtering algorithms \cite{bandyopadhyay_2018}, it does not combine likelihoods of data from different time instants. Instead, likelihoods are combined with \emph{time-adjusted} beliefs. These properties make DHS communication efficient and successful in tracking highly dynamic state transitions. Note that \cite{kayaalp2022dslw,kayaalp_dist_bayesian} deal with state estimation task only, and there are no rewards or actions in their setting. Therefore, we make proper modifications to the algorithm in the sequel.

In addition to these, the analysis in the current work is related to literature on the distributed optimization over networks \cite{sayed_2014,nedic2009distributed,dimakis2010gossip,di2016next}. In particular, we adopt the two-step approach from \cite{chen2015learning,chen2015learning2,kayaalp_dif_maml}. In the first step, these works establish that agents cluster around the network centroid, and then, they show that this centroid converges to a neighborhood of the optimal solution, under constant learning rates. However, their focus is on optimization and supervised learning rather than reinforcement learning, which creates non-trivial distinctions in the analysis. \\

\noindent \textbf{Notation.} Random variables are denoted in bold. For \( K \) vectors \( w_1 , w_2, \dots, w_K \in \mathbb{R}^M\) of dimension $M\times 1$ each, and for arbitrary matrices $\{A,B\}$, the notation \( \mcl \{w_k\}_{k=1}^K\) and \( \mdiag{\{A,B\}} \) stand for
\begin{equation}
    \mcl \{w_k\}_{k=1}^K = \begin{bmatrix}
    w_1 \\
    w_2 \\
    \vdots \\
    w_K
    \end{bmatrix}, \quad \mdiag{\{A,B\}} = \begin{bmatrix}
    A & 0 \\
    0 & B 
    \end{bmatrix}.
\end{equation}
The $\ell_p$-norm for a vector $w$ is represented by $\| w \|_p$, while the $\ell_p$-induced norm for a matrix $A$ is represented by $\| A \|_p$. To simplify the notation, we use $\| w \|$ and $\| A \|$ to denote the $\ell_2$-norm, without explicitly stating the subscript. The all-ones vector of dimension \( K \) is denoted by \( \mathds{1}_K  \). The symbol \( \otimes \) represents the Kronecker product. The Kullback-Leibler divergence \cite{csiszar11} between two distributions \( \mu_1 , \mu_2 \) is denoted by \( \dkl(\mu_1||\mu_2)\). We use the notation ``proportional to'', i.e., \( \propto \), whenever the LHS of the expression is the normalized version of the RHS. For example, for \( s \in \mathcal{S}\) and function \( f \):
\begin{align}
    \mu (s) \propto f (s) \Longleftrightarrow \mu (s) = \frac{f (s)}{\sum_{s^\prime \in \mathcal{S} }f (s^\prime)}.
\end{align}

\section{PRELIMINARIES}\label{sec:preliminaries}
In this work, we are interested in multi-agent policy evaluation under partially observable stochastic environments. For clarity of the exposition and to motivate the notation, we briefly review the procedure of single-agent policy evaluation under both fully and partially observable states.

\subsection{Fully-Observable Case}
For modeling a learning agent under fully observable and dynamic environments, the traditional setting is a finite Markov Decision Process (MDP). An MDP is defined by the quintuple \((\mathcal{S},\mathcal{A},\mathbb{T},\bm{r},\gamma) \), where \( \mathcal{S} \) is a set of states with cardinality \( | \mathcal{S} | = S\), \( \mathcal{A} \) is a set of actions, \( \mathbb{T} \) is a transition model where \( \mathbb{T} (s^\prime | a, s)\) denotes the probability of transitioning from \( s \in \mathcal{S} \) to \( s^\prime \in \mathcal{S}\) when the agent executes action \( a \in \mathcal{A} \), \( \bm{r} (s,a,s^\prime)\) denotes the reward the agent receives when it executes action \( a \) and the environment transitions from state \( s \) to \( s^\prime \), and \( \gamma \in [0,1) \) is a discount factor that determines the importance given to immediate rewards \( (\gamma \to 0) \) or the total reward \( (\gamma \to 1)\). 

The goal of policy evaluation is to learn the value function \( V^{\pi} (s) \) of a target policy \( \pi (a|s) \), where the value function is the expected return if the agent starts from state \( s \) and follows policy \( \pi \), namely,
\begin{equation}
    V^{\pi} (s) = \mathbb{E} \Big [ \sum_{i=0}^\infty \gamma^{i} \bm{r} (\bm{s}_{i},\bm{a}_i,\bm{s}_{i+1}) | \bm{s}_0 = s  \Big ],
\end{equation}
where \( \bm{s}_i \) is the state at time $i$ and \( \bm{a}_i \) is the action chosen by the agent according to the policy, \(\bm{a}_i \sim \pi (a|\bm{s}_i)\). In many applications, the state space is too large (or infinite), which makes it impractical to keep track of the value function for all states. Therefore, function approximations are used to reduce the dimension of the problem. For instance, linear approximations, which are the focus of the theoretical analysis of this work, correspond to using a parameter \( w^\circ~\in~\mathbb{R}^M \) to approximate \(V^{\pi}(s) \approx \phi (s)^{\T} w^\circ  \), where \( \phi: \mathcal{S} \to \mathbb{R}^M \) is a \emph{pre-defined} feature mapping for representing state \( s \).

A standard stochastic approximation algorithm to learn \( w^\circ \) from data is TD-learning \cite{sutton2018reinforcement,sayed_2022} such as the TD(0) strategy \cite{sutton1988learning} and variations thereof. If we denote the value function estimate at \( w \in \mathbb{R}^M\) by \( \widehat{V}(s,w) \triangleq \phi(s)^{\T} w\), then, under this strategy, the agent first computes the TD-error \( \bm{\delta}_i  \) at time \( i \) by using the observed transition tuple \( (\s_i, \rb_i , \s_{i+1}) \):
\begin{equation}\label{eq:delta_definition}
    \bde_i = \rb_i + \gamma \widehat{V} (\s_{i+1},\w_i) - \widehat{V} (\s_i,\w_i),
\end{equation}
where \( \rb_i \triangleq \rb (\s_i, \ac_i , \s_{i+1}) \) is the instantaneous reward at time \( i \). Subsequently, the agent uses this error to update the current parameter estimate \( \w_i \) to
\begin{equation}\label{eq:td0_no_reg}
    \w_{i+1} = \w_i + \alpha \bde_i \nabla_w \widehat{V} (\s_i,\w_i),
\end{equation}
where \( \alpha > 0 \) is the learning rate, and
\begin{align}
    \nabla_w \widehat{V} (\s_i,\w_i) = \phi (\s_i) 
\end{align}
for the linear function approximation case. This algorithm can be viewed as a ``stochastic gradient algorithm'' where the effective stochastic gradient is \( \gb_i \triangleq -\bde_i \phi (\s_i) \). In this work, we consider an \(\ell_2 \)-regularized version of the algorithm, which changes the update step \eqref{eq:td0_no_reg} to
\begin{align}\label{eq:td0_with_reg}
     \w_{i+1} = (1-2\rho\alpha)\w_i + \alpha \bde_i \nabla_w \widehat{V} (\s_i,\w_i),
\end{align}
where \( \rho > 0 \) is a constant hyper-parameter. As opposed to supervised learning, regularization is rather under-explored in reinforcement learning, with notable exceptions in \cite{kolter2009regularization,hoffman2011regularized}. However, recent work \cite{farebrother2018generalization,cobbe19quantifying} suggests that regularization can increase generalization and sample-efficiency in function approximation with over-parameterized models. 

\subsection{Partially-Observable Case}\label{sec:single_pomdp}

In many applications, the agent does not directly observe the state \( \s_i\). For instance, a robot may receive noisy and partially informative observations from its sensors about the environment. The observation \( \bxi_i\) that the agent receives at time \( i \) is generally assumed to be distributed according to some likelihood function linking it to the unobservable state, say, \( \bxi_i \sim L(\xi| \s_i)\), which is conditioned on \( \s_i \). In these scenarios, the agent will need to estimate the latent state first from the observations. To do so, the agent will need to learn a probability vector \( \bmu_i \in \mathcal{M}(S) \) over the set of states \( \mathcal{S} \), which is called the \emph{belief} vector \cite{sondik1978optimal,sayed_2022}. Here, \( \mathcal{M}(S) \) denotes the \( S\)-dimensional probability simplex, and the entry \( \bmu_i (s) \in [0,1]\) of the belief vector quantifies the confidence the agent has about state \( s \) being the true state at time \( i \). The value of $\bm{\mu}_i(s)$ corresponds to the posterior probability of $s$ conditioned on the action-observation history (a.k.a. trajectory):
\begin{equation}\label{eq:fi_definition}
     \bmf_i \triangleq \{ \bxi_i, \ac_{i-1}, \bxi_{i-1}, \dots \},
\end{equation}
 which means:
\begin{equation}\label{eq:true_posterior}
\bmu_i (s) \triangleq \mathbb{P} (\s_i = s | \bmf_i).
\end{equation}
This posterior satisfies the following temporal recursion \cite{sondik1978optimal,sayed_2022,krishnamurthy_2016}:
\begin{equation}\label{eq:centralized_posterior}
 \bmu_i (s) \propto L(\bxi_i|s) \bmeta_i (s),
\end{equation}
where \( \bmeta_i (s) \) is the time-adjusted prior defined by
\begin{equation}\label{eq:ck_cent}
\bmeta_i (s)\!\triangleq \! \mathbb{P} (\s_i = s | \bmf_{i-1}^a)\! =\!\!\!\sum_{s^\prime\in\mathcal{S}} \bT(s| s^\prime,\ac_{i-1}) \bmu_{i-1}(s^\prime).
\end{equation}
Here, $\bmf_{i-1}^a$ is the collection of past observations and actions, i.e.,
\begin{equation}\label{eq:fia_definition}
    \bmf_{i-1}^a  \triangleq \{ \ac_{i-1}, \bxi_{i-1},\ac_{i-2}, \dots \},
\end{equation}
where it is important to notice that $\bmf_i = \{ \bxi_i\} \cup \bmf_{i-1}^a$. If beliefs are used as substitutes for hidden states, then partially-observable MDPs (POMDPs) can be treated as \emph{continuous} MDPs, since beliefs are continuous even if the number of states is finite. In this way, the policy evaluation problem would correspond to evaluating \( V^{\pi} (\mu) \) where the value function is now defined as the expected return when the agent starts from the belief state \( \mu \) and follows the policy \( \pi(a|\mu)\), namely \cite{sondik1978optimal,sayed_2022}:
\begin{equation}
    V^{\pi} (\mu) = \mathbb{E} \Big [ \sum_{i=0}^\infty \gamma^{i} \bm{r}_i| \bmu_0 = \mu  \Big ].
\end{equation}
Observe that, in contrast to the fully-observable case, the agent now chooses action \( \ac_i \) according to the policy \(\ac_i~\sim~\pi(a|\bmu_i) \), which is conditioned on the belief vector.

Algorithm \eqref{eq:delta_definition}--\eqref{eq:td0_with_reg} can be adjusted for POMDPs by using the belief vectors $(\bm{\mu}_i,\bm{\eta}_{i+1})$ instead of the states $(\s_i,\s_{i+1})$. Thus, we let 
\begin{equation}\label{eq:delta_pomdp}
   \bde_i = \rb_i + \gamma \widehat{V} (\bmeta_{i+1},\w_i) - \widehat{V} (\bmu_i,\w_i),
\end{equation}
and
\begin{equation}\label{eq:w_update_pomdp}
     \w_{i+1} = (1-2\rho\alpha)\w_i + \alpha \bde_i \nabla_w \widehat{V} (\bmu_i,\w_i),
\end{equation}

where the approximations $\widehat{V}(\mu,w)$ are computed by using the feature vectors $\phi(\mu)$, now dependent on $\mu$, to evaluate \( \widehat{V}(\mu,w)~\triangleq~\phi(\mu)^{\T} w \). Note that from now on \( \phi: \mathcal{M}(S) \to \mathbb{R}^M \) is a different feature mapping that represents $\mu$ instead of $s$, and agents' goal is to learn $w^\circ$ that satisfies $V^{\pi}(\mu)~\approx~\phi~(\mu)^{\T}~w^\circ$.

Observe from \eqref{eq:centralized_posterior}--\eqref{eq:ck_cent} that in order for the agent to update the belief vectors $(\bm{\mu}_i,\bm{\eta}_{i+1})$, it needs to know the transition model \( \mathbb{T} \) and the likelihood functions \( L(\bxi_i|s) \) for each state. However, the agent does not need to know the underlying reward model \( \rb\). It can use instantaneous reward samples \( \rb_i \) to run the algorithm. In this sense, the algorithm is a mixture of model-based and model-free reinforcement learning. Motivation for this approach is at least two-fold. First, in some applications, learning the transition and observation models from data is inherently easier than learning the reward function. This is because the reward function can depend on some latent characteristics of the environment or some human expert, which may be challenging to estimate. One example where this scenario can arise is autonomous cars \cite{kiran2022}. In this case, the observations from environmental sensors and cameras are processed with a learned likelihood model such as a convolutional neural network. The transition dynamics of the car depends on various parameters such as speed, acceleration, position, and incline, and can be modeled based on physics laws and mapping of the surroundings. However, learning a reward function for this application is notoriously difficult, as it is challenging to cover all possible situations \cite{biyik2022learning}. Second, the agent can still run \eqref{eq:delta_pomdp}--\eqref{eq:w_update_pomdp} even if beliefs are not formed through \eqref{eq:centralized_posterior}--\eqref{eq:ck_cent}, but estimated by some other approach, as in \cite{moreno2021neural,muglich2022,mao2020}. 
 
\section{MULTI-AGENT POLICY EVALUATION}\label{sec:multi_agent}

We now consider a set \( \mathcal{K} \) of \( K \) cooperative agents that aim to evaluate the average value function under a joint policy \( \pi = \{ \pi_k\}_{k=1}^K \) that consists of individual policies \( \pi_k\). The framework we consider is a \emph{decentralized} POMDP (Dec-POMDP) \cite{oliehoek2016concise}, which is defined by the sextuple \( (\mathcal{S},\mathcal{A}_k,\mathcal{O}_k,\mathbb{T},\bm{r}_k,\gamma) \). Here, the set of states \( \mathcal{S} \) and the transition model \( \mathbb{T} \) are common to all agents, where the notation \( \mathbb{T} (s|s^\prime,a) \) now specifies the probability that the environment transitions from \( s^\prime\) to \( s\) when the agents execute the joint action \(a= \{a_k\}_{k=1}^K \). The individual action \( a_k \) of each agent \( k \) takes values from the set \( \mathcal{A}_k \), and \( \bm{r}_k (s,a,s^\prime)\) is the \emph{local} reward \( k \) gets when the agents execute the collection of actions \(a \) and the environment transitions from \( s \) to \( s^\prime \). Note that this setting covers general teamwork scenarios where the local reward of an individual agent can be dependent on all actions, and not only on its own actions. Specifically, it covers the scenarios that all agents observe the same reward, i.e., \( \bm{r}_k (s,a,s^\prime) = \bm{r} (s,a,s^\prime), \forall k \in \mathcal{K}\). Remember that agents receive instantaneous rewards as they progress through the POMDP, and they are not required to know the joint action $a$ from all agents. Moreover, \( \mathcal{O}_k \) is a set of \emph{private} observations. At each time instant \( i \), agent \( k \) receives observation \( \bxi_{k,i} \in \mathcal{O}_k \) emitted by state \( \s_i\), and assumed to be distributed according to the local \emph{marginal} likelihood \( L_k(\xi_k| \s_i) \).

Similar to the single-agent case, Dec-POMDPs can be treated as multi-agent belief MDPs by replacing the hidden states with joint centralized beliefs defined by \cite[Chap. 2]{oliehoek2016concise}
\begin{equation}\label{eq:centralized_posterior_ma_def}
 \bmu_i (s) \triangleq \mathbb{P} (\s_i = s | \bmf_i) \propto L(\bxi_i|s) \bmeta_i (s).
\end{equation}
Here, \(\bmf_{i} \) denotes the history of all observations and past actions from across all agents until time \( i \), where in the definition \eqref{eq:fi_definition}, \( \bxi_i \triangleq \{\bxi_{k,i}\}_{k=1}^K\) is now the aggregate of the observations from across the network, and \( \ac_{i-1}\) is a tuple aggregating actions from all agents at time \( i-1\). Moreover, under spatial independence, the joint likelihood \( L(\bxi_i|s) \) appearing in \eqref{eq:centralized_posterior_ma_def} is given by 
\begin{equation}
    L(\bxi_i|s) = \prod_{k=1}^K L_k(\bxi_{k,i}| s).
\end{equation}
In a manner similar to the single-agent case, the belief \( \bmeta_i (s) \) is the time-adjusted prior conditioned on $\bmf_{i-1}^a$ \eqref{eq:fia_definition}:
\begin{equation}\label{eq:ck_cent_ma_def}
\bmeta_i (s) \!\triangleq\!  \mathbb{P} (\s_i = s | \bmf_{i-1}^a)\! =\!\!\sum_{s^\prime\in\mathcal{S}} \bT(s| s^\prime, \ac_{i-1}) \bmu_{i-1}(s^\prime).
\end{equation}
The goal of policy evaluation is to learn the \emph{team} value function, which is the expected average reward of all agents starting from some belief state \( \mu \), i.e.,
\begin{equation}\label{eq:true_value_func}
    V^{\pi} (\mu) = \mathbb{E} \Big [ \sum_{i=0}^\infty \gamma^{i}  \Big ( \frac{1}{K} \sum_{k=1}^K \bm{r}_{k,i}  \Big ) \Big | \bmu_0 = \mu  \Big ],
\end{equation}
where \( \bm{r}_{k,i} \) denotes the instantaneous local reward agent \( k \) gets at time \( i \).

There is one major inconvenience with this approach. In order to compute the joint belief \eqref{eq:centralized_posterior_ma_def}, it is necessary to fuse all observations and actions from across the agents in a central location. This is possible in settings where there exists a fusion center. However, many applications rely solely on localized processing. In the following, we discuss and compare two strategies for multi-agent reinforcement learning under partial observations: \((i)\) a centralized strategy, \((ii)\) and a fully decentralized strategy.

\subsection{Centralized Strategy}\label{sec:centralized_case}

In the fully centralized strategy, the state estimation and policy evaluation phases are centralized and, hence, the setting is equivalent to a single-agent POMDP, already discussed in Sec.~\ref{sec:preliminaries}-\ref{sec:single_pomdp}, using the joint likelihood \(L(\bxi_i|s)\) and the average reward \( \rb_i \triangleq K^{-1} \sum_{k=1}^K \bm{r}_{k,i} \). The fusion center computes the joint belief \eqref{eq:centralized_posterior_ma_def}, and agents take actions based on this joint belief, i.e., \(  \ac_{k,i} \sim \pi_k (a_k|\bmu_{i}) \). The fusion center then computes the centralized TD-error:
\begin{equation}\label{eq:cent_pol_learn_del_def}
   \bde_i = \rb_i + \gamma \widehat{V} (\bmeta_{i+1},\w_i) - \widehat{V} (\bmu_i,\w_i),
\end{equation}
and updates the estimate to
\begin{equation}\label{eq:cent_pol_learn_w_def}
     \w_{i+1} = (1-2\rho\alpha)\w_i + \alpha \bde_i \nabla_w \widehat{V} (\bmu_i,\w_i).
\end{equation}
This construction is listed under Algorithm~\ref{alg:fully_cent}. 

  \begin{algorithm}[]
 \caption{Centralized policy evaluation under POMDPs}
 \begin{algorithmic}[1]
    \item set initial prior $\eta_{0}(s)>0$,  $\forall s \in\mathcal{S}$
  \item initialize $w_{0}$
    \While{$i\geq 0$} 
  \State each agent $k$ observes $\bxi_{k,i}$ 
\State collect all observations \( \bxi_i \triangleq \{\bxi_{k,i}\}_{k=1}^K\) and \textbf{evaluate}
\begin{equation}\label{eq:centralized_posterior_ma}
 \bmu_i (s) \propto L(\bxi_i|s) \bmeta_i (s)
\end{equation}
 \For{each agent $k\in\mathcal{K}$}
\State Take action \( \ac_{k,i} \sim \pi_k (a_k|\bmu_{i}) \)
\State Get reward  \( \rb_{k,i} = \bm{r}_k (\s_i,\ac_{i},\s_{i+1}) \)
\EndFor
\State then, \textbf{evolve}
\begin{equation}\label{eq:ck_cent_ma}
\bmeta_{i+1} (s) = \sum_{s^\prime\in\mathcal{S}} \bT(s| s^\prime,\ac_{i}) \bmu_{i}(s^\prime)
\end{equation}
 \State \textbf{average} the rewards  \( \rb_i = \frac{1}{K} \sum_{k=1}^K \bm{r}_{k,i} \)
 \State \textbf{update} the model:
\State \begin{equation}\label{eq:cent_pol_learn_del}
  \bde_i = \rb_i + \gamma \widehat{V} (\bmeta_{i+1},\w_i) - \widehat{V} (\bmu_i,\w_i)
\end{equation}
\begin{equation}\label{eq:cent_pol_learn_w}
     \w_{i+1} = (1-2\rho\alpha)\w_i + \alpha \bde_i \nabla_w \widehat{V} (\bmu_i,\w_i)
\end{equation}
\EndWhile
 \end{algorithmic} \label{alg:fully_cent}
 \end{algorithm}

\subsection{Decentralized Strategy}\label{sec:network_learning}

The centralized strategy is disadvantageous in the sense that $(i)$ failure of the fusion center results in failure of the system; $(ii)$ there can be communication bottlenecks at the fusion center; $(iii)$ and agents can be spatially distributed to begin with. Therefore, in this section, we propose a fully decentralized strategy for policy evaluation where agents communicate with their immediate neighbors only.

\subsubsection{Decentralized Network Model}

\begin{figure}
   \centering
    \includegraphics[width=0.6\textwidth]{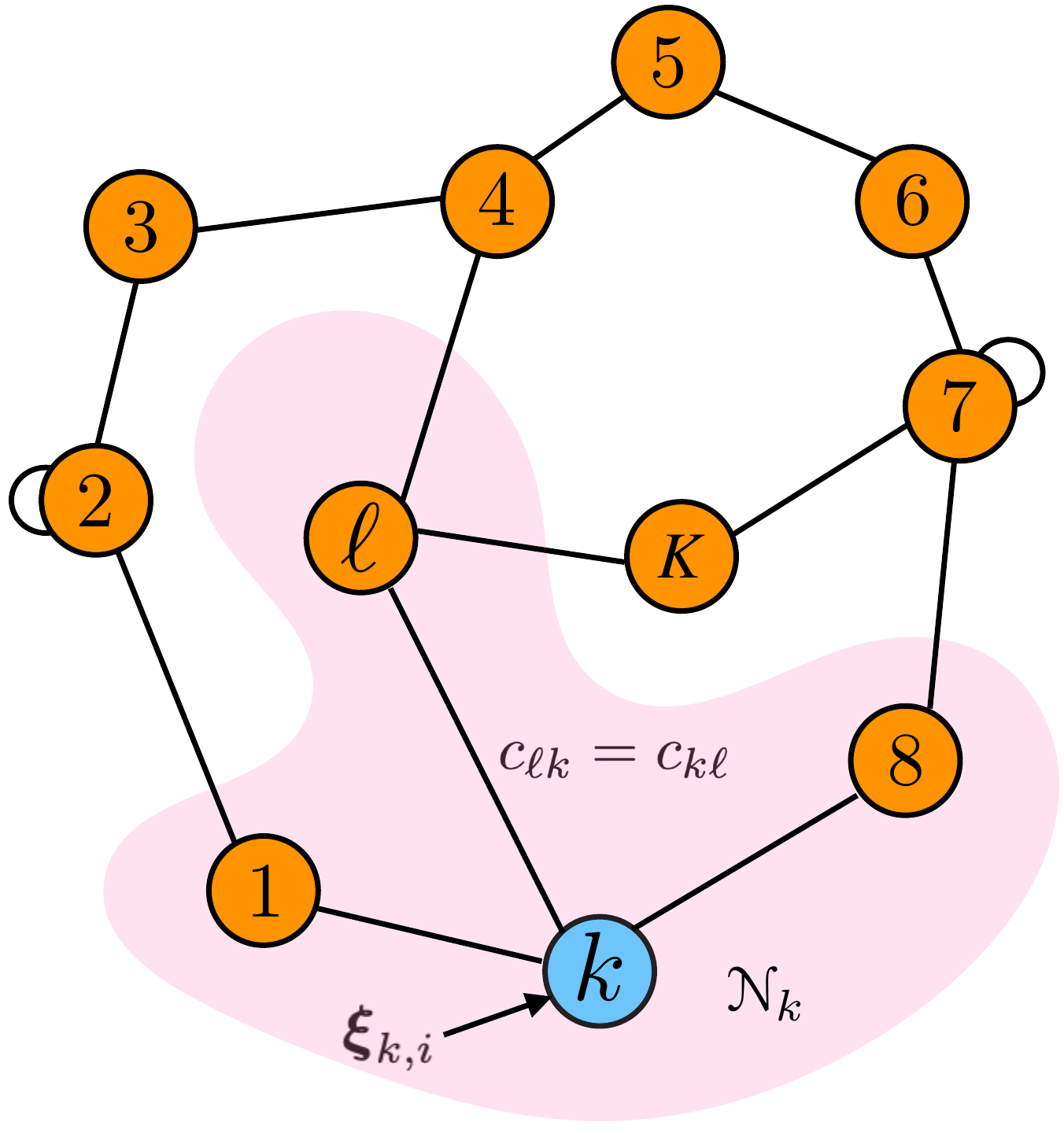}
    \caption{An illustration of a network model.}
   \label{fig:network_model}
    \end{figure}

We refer to Fig.~\ref{fig:network_model} and assume that the graph is strongly connected \cite{sayed_2014}, which means that paths exist connecting any pair of agents \( (\ell,k)\) in both directions, and in addition, there exists at least one agent in the graph that does not discard its own information (i.e., $c_{kk}>0$ for at least one agent $k$). Under this assumption, the combination matrix \( C = [c_{\ell k}]\), where entry \( c_{\ell k} \geq 0 \) scales the information agent \( k \) receives from agent \( \ell\), becomes primitive. If two agents are not connected by an edge then $c_{\ell k}=0$. We assume $C$ is symmetric and doubly-stochastic, meaning that 
\begin{equation}
    \sum_{\ell = 1}^K c_{\ell k} = 1, \quad c_{\ell k} = c_{k \ell},
\end{equation}
or in matrix notation:
\begin{equation}\label{eq:c_matrix_assumption}
C\mathds{1}_{K}=\mathds{1}_{K}, \quad C=C^{\T}.
\end{equation}

\subsubsection{Local Belief Formation}

In the fully decentralized strategy, the agents cannot form the joint belief \eqref{eq:centralized_posterior_ma_def} since they do not have access to the observations and actions of all other agents. They, however, can construct local beliefs. To do so, we will extend the diffusion HMM strategy (DHS) from \cite{kayaalp2022dslw} and \cite{kayaalp_dist_bayesian}, which is originally designed for hidden Markov models, to the current POMDP setting. 

In DHS, the global belief vectors $\{\bm{\mu}_i, \bm{\eta}_i\}$ are replaced by local belief vectors $\{\bm{\mu}_{k,i},\bm{\eta}_{k,i}\}$, and the latter are updated by using local observations and by relying solely on interactions with the immediate neighbors. The original DHS algorithm is designed for actionless partially observable Markov chains, and each agent can use the same global transition model. However, in POMDPs, transition of the global state depends on the joint action, and the agents cannot perform a \emph{centralized} time-adjustment step as in \eqref{eq:ck_cent_ma} since they do not know the actions of all agents in the network.

Therefore, one strategy is to use a transition model that is obtained by marginalizing over actions that are unknown to agent $k$. More specifically, let $a_{{\cal N}_k} \in \mathcal{A}_{{\cal N}_k}$ denote a tuple of actions taken by the set of neighbors of agent $k$ (which we are denoting by ${\cal N}_k$). These actions can be assumed to be known by agent $k$ if, for instance, agents share their actions with their neighbors. Let  $a_{{\cal N}_k}^c \in \mathcal{A}_{{\cal N}_k}^c$  denote the remaining actions by all other agents in the network, so that $a=a_{{\cal N}_k}\cup a_{{\cal N}_k}^c$. Then, each agent can use the following \emph{local} transition model approximation: 
\begin{equation}\label{eq:approximate_trans_prof}
     \bT^\pi_k(s| s^\prime, a_{{\cal N}_k})  \propto \!\!\!\!\!\!\!\! \sum_{a_{{\cal N}_k}^c \in  \mathcal{A}_{{\cal N}_k}^c} \!\!\! \!\!\!\!\bT(s| s^\prime, a_{{\cal N}_k}, a_{{\cal N}_k}^c ) \pi (a_{{\cal N}_k},  a_{{\cal N}_k}^c | s^\prime)
\end{equation}

in lieu of \( \mathbb{T} (s|s^\prime,a) \), to time-adjust its local belief:
\begin{align}\label{eq:dif_evolve_step_1}
  \bmeta_{k,i} (s) = \sum_{s^\prime \in \mathcal{S}} \bT^\pi_k(s| s^\prime, \ac_{{\cal N}_k,i-1}) \bmu_{k,i-1}(s^\prime),
\end{align}
Here, $\ac_{{\cal N}_k,i-1}$ is the tuple of actions taken by the neighbors of agent $k$ at time instant $i-1$. Moreover, in \eqref{eq:approximate_trans_prof}, the notation \(\pi (a_{{\cal N}_k},  a_{{\cal N}_k}^c | s^\prime)\) represents the joint action probability:
\begin{equation}
    \pi (a_{{\cal N}_k},  a_{{\cal N}_k}^c | s^\prime) = \prod_{\ell=1}^K \pi_{\ell} (a_{\ell} | s^\prime),
\end{equation}
where the notation \( \pi (a| s) \) is now a shorthand for \( \pi (a| \mu )\) when 
\begin{equation}
    \mu  = [0  \dots 1 \dots 0]^{\T},
\end{equation}
i.e., when the belief attains value 1 for state \( s \) and is 0 otherwise. Note that this construction leads to a richer scenario compared to \cite{kayaalp2022dslw,kayaalp_dist_bayesian}, with transition models that are different across the agents.

The rest of the algorithm is the same as the DHS strategy. Following \eqref{eq:dif_evolve_step_1}, and based on the personal observation \( \bxi_{k,i} \), each agent $k$ forms an \emph{intermediate} belief using a \(\beta\)-scaled Bayesian update of the form:
 \begin{align}\label{eq:dif_adapt_step_1}
    \bpsi_{k,i} (s) \propto (L_k(\bxi_{k,i} | s))^{\beta}\bmeta_{k,i} (s) ,
\end{align}

where \( \beta > 0\). Finally, agents in the neighborhood of $k$ share their intermediate beliefs, which allows agent $k$ to update its belief using the weighted geometric average expression:
\begin{align}\label{eq:dif_combine_step_1}
    \bmu_{k,i}(s) \propto  \prod_{\ell \in \mathcal{N}_k}  \big ( \bpsi_{\ell,i} (s) \big )^{c_{\ell k}}  .
\end{align}

This procedure of repeated updating and exchanging of beliefs allows information to diffuse over the network.

\subsubsection{Diffusion Policy Evaluation}
In the fully decentralized strategy, the local belief formation strategy is used during both training and execution phases. Namely, the target value function in \eqref{eq:true_value_func} represents the average return agents get when they execute the policy \(\pi\) with their local beliefs formed via the DHS strategy. Moreover, since the policy evaluation is also decentralized, during the training phase, they again need to use DHS to approximate the global belief state \(\mu\) on top of the function approximation. More specifically, using its local belief vectors, each agent $k$ computes a local TD error:
\begin{equation}\label{eq:delta_parameter_def}
   \bde_{k,i} = \rb_{k,i} + \gamma \widehat{V} (\bmeta_{k,i+1},\w_{k,i}) - \widehat{V} (\bmu_{k,i},\w_{k,i}),
\end{equation}
where \( \rb_{k,i} = \bm{r}_k (\s_i,\ac_{i},\s_{i+1}) \) is also a function of the local beliefs since each agent \( k \) now executes the action \( \ac_{k,i} \sim \pi_k (a_k | \bmu_{k,i}) \). Subsequently, each agent $k$ forms an intermediate parameter estimate denoted by 
\begin{equation}\label{eq:adapt_parameter_def}
     \bz_{k,i+1} = (1-2\rho\alpha)\w_{k,i} + \alpha \bde_{k,i} \nabla_w \widehat{V} (\bmu_{k,i},\w_{k,i}).
\end{equation}
After receiving the intermediate estimates from its neighbors, agent $k$ updates $\w_{k,i}$ to 
\begin{equation}\label{eq:combine_parameter_def}
    \w_{k,i+1} = \sum_{\ell \in \mathcal{N}_k} c_{\ell k} \bz_{\ell,i+1}.
\end{equation}
The local adaptation step \eqref{eq:adapt_parameter_def} followed by the combination step \eqref{eq:combine_parameter_def} are reminiscent of diffusion strategies for distributed learning \cite{sayed_2014,sayed_2022}. Observe that there are actually two combination steps involved in diffusion policy evaluation: the belief combination \eqref{eq:dif_combine_step_1} with geometric averaging (GA), and the parameter combination \eqref{eq:combine_parameter_def} with arithmetic averaging (AA). These choices of fusion rules are supported by recent results in the literature \cite{li2019aaga,kayaalp2022arithmetic} that promote the use of GA for probability density functions and AA for point estimates. The listing of the proposed diffusion policy evaluation strategy for POMDPs appears in Algorithm~\ref{alg:fully_decent}.
  \begin{algorithm}[]
 \caption{Diffusion policy evaluation under POMDPs}
 \begin{algorithmic}[1]
  \item set initial priors $\eta_{k,0}(s)>0$,  $\forall s \in\mathcal{S}$ and  $\forall k\in \mathcal{K}$ 
  \item choose $\beta>0$
  \item initialize $w_{k,0}$ for $\forall k\in \mathcal{K}$ 
    \While{$i\geq 0$} 
  \State each agent $k$ observes $\bxi_{k,i}$ 
 \FORC{each agent $k\in\mathcal{K}$ and $s \in \mathcal{S}$}
 \begin{align}\label{eq:dif_adapt_step}
    \bpsi_{k,i} (s) \propto (L_k(\bxi_{k,i} | s))^{\beta}\bmeta_{k,i} (s)  
\end{align}
\begin{equation}\label{eq:dif_combine_step}
    \bmu_{k,i}(s) \propto \prod_{\ell \in \mathcal{N}_k}  \big ( \bpsi_{\ell,i} (s) \big )^{c_{\ell k}}    
\end{equation}
\ENDFORC
 \For{each agent $k\in\mathcal{K}$}
\State Take action \( \ac_{k,i} \sim \pi_k (a_k|\bmu_{k,i}) \)
\State Get reward  \( \rb_{k,i} = \bm{r}_k (\s_i,\ac_{i},\s_{i+1}) \)
\EndFor
\FORB{each agent $k\in\mathcal{K}$}
\State Compute \( \bT^\pi_k(s| s^\prime, \ac_{{\cal N}_k,i}) \) using \eqref{eq:approximate_trans_prof}, and
\begin{equation}\label{eq:dif_evolve_step}
  \bmeta_{k,i+1} (s) = \sum_{s^\prime \in \mathcal{S}} \bT^\pi_k(s| s^\prime, \ac_{{\cal N}_k,i}) \bmu_{k,i}(s^\prime)
\end{equation}
\ENDFORB
 \FORF{each agent $k\in\mathcal{K}$}
\begin{equation}\label{eq:delta_parameter}
  \bde_{k,i} = \rb_{k,i} + \gamma \widehat{V} (\bmeta_{k,i+1},\w_{k,i}) - \widehat{V} (\bmu_{k,i},\w_{k,i})
\end{equation}
\begin{equation}\label{eq:adapt_parameter}
     \bz_{k,i+1} = (1-2\rho\alpha)\w_{k,i} + \alpha \bde_{k,i} \nabla_w \widehat{V} (\bmu_{k,i},\w_{k,i})
\end{equation}
\ENDFORF
 \FORE{each agent $k\in\mathcal{K}$}
\begin{equation}\label{eq:combine_parameter}
    \w_{k,i+1} = \sum_{\ell \in \mathcal{N}_k} c_{\ell k} \bz_{\ell,i+1}
\end{equation}
\ENDFORE
\EndWhile
 \end{algorithmic} \label{alg:fully_decent}
 \end{algorithm}
 
Algorithm~\ref{alg:fully_decent} has the following listed advantages:
 \begin{itemize}
     \item \textbf{Decentralized information structure:} The algorithm is designed to be fully decentralized, with each agent only having access to its own private data, such as observations and rewards, without the need to share this information with other agents. Importantly, agents do not require knowledge of the joint distribution of observations or the network topology. They only know their own marginal likelihood function, and their actions are only known by (or transmitted to) their immediate neighbors. If agents happen to know their own marginal transition models, they do not need to know the policies of other agents or the global transition model. However, if the application requires them to approximate it themselves, they require knowledge of the other policies and the global transition model. 
     \item \textbf{Privacy}: The algorithm is also advantageous in terms of privacy since \((i)\) communicating beliefs allows information diffusion without explicitly sharing raw observational data, and \((ii)\) exchanging value parameters allows agents learn the cumulative reward across network without explicitly sharing local rewards.
     \item \textbf{Complexity}: \((i)\) The memory requirement is constant over time, with each agent only needing to store its value function parameter estimate ($M$-dimensional) and local belief ($S$-dimensional), as well as the necessary model functions. \((ii)\) The communication requirement is also manageable, with each agent communicating only with its immediate neighbors through belief and parameter sharing. The communication load is not affected by the network size, making our algorithm scalable and avoiding communication bottlenecks. \((iii)\) The computational complexity depends on whether the application at hand allows agents to have access to the local transition model. If this is the case, then the computational complexity is equivalent to the single-agent Bayesian filtering case, which is $O(S^2)$. The combination steps add only linear additional complexity $O(S)$ with fixed neighborhood size. However, if agents need to approximate the transition model themselves, the computational complexity increases with the network size, and becomes $O(K S^2)$. This is due to the need to average over non-neighbors' actions in \eqref{eq:approximate_trans_prof}, whose size grows with the network size in general. Compared to alternative approaches such as relaying raw data, incremental approaches \cite{bertsekas2015parallel}, or Bayesian belief forming \cite{acemoglu_2011}, our algorithm is much lighter in terms of complexity. Relaying raw data, for example, would result in an exponential increase of memory and communication overload at each hop, making it highly impractical.
     The incremental approach of relaying over a cyclic path (which is NP-hard to find \cite{garey1979computers}) that visits each agent once would reduce the overload. However, it is not robust against failures and not scalable, making it impractical for a decentralized setting. The Bayesian belief forming strategy requires knowledge of the network topology and other agents' functions, and known to be NP-hard, even in the much simpler case of fixed state and no action setting \cite{hazla2021bayesian}. 
 \end{itemize}

\section{MAIN RESULTS}\label{sec:theoretical_results}

In this section, we analyze the performance of the decentralized strategy in Algorithm~\ref{alg:fully_decent}. In particular, we first show in Sec.~\ref{sec:network_disagreement} that the value function parameters $\{\w_{k,i}\}$ of the agents cluster around the network centroid. Then, in Sec.~\ref{sec:centralized_disagreement}, we show that this network centroid has a bounded difference from the parameter of a baseline strategy (which will be presented in Algorithm~\ref{alg:cent_dec_baseline}). Our analysis relies on bounding the disagreement between the joint centralized belief $\bmu_{i}$ and the local estimate $\bmu_{k,i}$, which is presented next.

\subsection{Belief Disagreement}

In a manner similar to \cite{kayaalp_dist_bayesian}, we introduce the following risk functions in order to assess the disagreement between the local beliefs formed via \eqref{eq:dif_adapt_step}--\eqref{eq:dif_evolve_step} with the joint centralized beliefs formed via \eqref{eq:centralized_posterior_ma}--\eqref{eq:ck_cent_ma}:
\begin{align}\label{eq:filter_risk}
J_{k,i}&\triangleq\e_{\f_i} \dkl(\bmu_i || \bmu_{k,i}) ,
\end{align}
and
\begin{align}\label{eq:prior_risk}
\widetilde{J}_{k,i}&\triangleq\e_{\f_{i-1}^a} \dkl(\bmeta_i || \bmeta_{k,i}) .
\end{align}
The risks in \eqref{eq:filter_risk} and \eqref{eq:prior_risk} measure the disagreement after and before the joint observation \( \bxi_i\), respectively. Remember that \cite{kayaalp_dist_bayesian} considers a naive state estimation setting rather than a POMDP. Specifically, in their setting, the transition model does not depend on actions, and it is assumed that every agent knows the global transition model accurately. In comparison, in the current work, each agent uses a local approximation for the global transition model based on \eqref{eq:approximate_trans_prof}. Therefore, we need to make some non-trivial adjustments to the belief disagreement analysis. We begin with adjusting the assumptions from \cite{kayaalp_dist_bayesian} to our model.
 \subsubsection{Modeling Conditions}
 \begin{itemize}
     \item \textbf{Likelihood functions}: Each observation has bounded information about the true state. More formally, 
\begin{equation}
    \dkl (L_k (\xi | s) || L_k (\xi | s^\prime)) < \infty
\end{equation}
     which ensures that likelihoods for each state pair ($s$, $s^\prime$) share the same support, and in addition to this,
 \begin{align}\label{eq:log_bound_assumption}
  \big |\log L_k (\xi | s) \big | \leq B
\end{align}
over its support for each state \( s \in \mathcal{S} \) and agent \( k \in \mathcal{K}\).
\item \textbf{Transition model}: The Markov chain induced by any joint action $a \in \mathcal{A}$ is \emph{irreducible} and \emph{aperiodic}. Since the number of states is finite, this assumption implies that the transition model \( \bT(s|s^\prime,a)\) is ergodic \cite[Chap. 2]{resnick2002}. Like \cite{kayaalp_dist_bayesian}, we focus on the important class of \emph{geometrically} ergodic models, which additionally satisfy the relation \( \kappa (\bT^a) \leq \kappa (\bT) \) for some constant $\kappa (\bT) < 1$. Here, \( \kappa (\bT^a) \) is the Dobrushin coefficient \cite[Chap. 2]{krishnamurthy_2016} defined by:
\begin{align}\label{eq:definition_dobrushin_coefficient}
    \kappa (\bT^a) \triangleq \sup_{s^\prime,s^{\prime\prime} \in \mathcal{S}} \frac{1}{2} \sum_{s \in \mathcal{S}} \big | T_{s s^\prime}^a - T_{s s^{\prime\prime}}^a \big | ,
\end{align}
where \( T_{s s^\prime}^a \triangleq \bT(s|s^\prime,a) \) is a generic entry of the $S \times S$ transition matrix $T^a$. Due to space limitations, we refer the reader to \cite[Chap. 2]{krishnamurthy_2016} for a comprehensive discussion on the Dobrushin coefficient \( \kappa(\bT^a)\). In short, \( \kappa (\bT^a) \) quantifies how fast the transition model forgets its initial conditions. Namely, as \(\kappa (\bT^a) \to 0 \), past conditions are forgotten faster. Instances of geometrically ergodic transition models include transition matrices with all positive elements, or that satisfy the minorization condition in \cite[Theorem 2.7.4]{krishnamurthy_2016}. In addition to this condition from \cite{kayaalp2022dslw,kayaalp_dist_bayesian}, we have an additional assumption on the transition model to regulate the disagreement stemming from the local transition model estimates:
\begin{assumption}[\textbf{Transition model disagreement}]\label{assumption:trans_disagreement} 
For each agent \(k\), consider the $n$-hop neighbors set $\mathcal{N}_{k^n}$ and its complement $\mathcal{N}_{k^n}^c$. In other words, $\mathcal{N}_{k^n}$ is the set of agents that have at most $n$-hop distance to the agent $k$. We define the transition model approximation that uses \(n\)-hop neighbors' actions as follows:
\begin{align}
 &\bT^\pi_k(s| s^\prime, a_{{\cal N}_{k^n}}) \notag \\ &\quad \qquad \propto \!\!\!\!\!\!\!\!\!\! \sum_{a_{{\cal N}_{k^n}}^c \in  \mathcal{A}_{{\cal N}_{k^n}}^c} \!\!\!\!\! \!\!\!\!\bT(s| s^\prime, a_{{\cal N}_{k^n}}, a_{{\cal N}_{k^n}}^c ) \pi (a_{{\cal N}_{k^n}},  a_{{\cal N}_{k^n}}^c | s^\prime)  .  
\end{align}
Then, we assume that 
\begin{equation}
    \dkl \Bigg (\bT^\pi_k \Big(s \big| s^\prime, a_{ {\cal N}_{\scriptstyle k^{\scriptstyle n}}} \Big ) \bigg | \bigg | \bT^\pi_k \Big(s \big | s^\prime, a_{{\cal N}_{\scriptstyle k^{\scriptstyle n+1}}} \Big) \Bigg ) < \infty,
\end{equation}
which ensures that transition model approximations induced from $n$-hop and $(n+1)$-hop neighbors' actions share the same support. Moreover, we assume that over the shared support,
\begin{equation}
    \Bigg |\log \frac{\bT^\pi_k \Big(s| s^\prime, a_{ {\cal N}_{\scriptstyle k^{\scriptstyle n}}} \Big)}{\bT^\pi_k \Big(s| s^\prime, a_{{\cal N}_{\scriptstyle k^{\scriptstyle n+1}}} \Big)} \Bigg| \leq \tau.
\end{equation}
for $n \geq 1$.
\end{assumption}$\qed$

This assumption basically makes sure that the increase in the error of the transition model approximation of agents due to lack of information about actions is bounded at each geodesic distance increase to that agent.

 \end{itemize}
 
 \subsubsection{Difference with Centralized Strategy}
 
The following result provides upper bounds on the disagreement measures in \eqref{eq:filter_risk}--\eqref{eq:prior_risk}.

 \begin{theorem}[{\bf Bounds on belief disagreement}]\label{th:kl_without_net_assumption}For each agent \( k \), the belief disagreement risks \eqref{eq:filter_risk} and \eqref{eq:prior_risk} get bounded with a linear rate of $\kappa (\bT)$. Namely,  as \( i \to \infty \),
\begin{align}\label{eq:th1_risk}
 J_{k,i} \leq   \frac{2 \sqrt{K} \beta  \lambda B}{1-\kappa (\bT)}+\frac{(K-d_{\textup{min}})\: \tau}{1-\kappa (\bT)}  
\end{align}
and
\begin{align}\label{eq:th1_prior}
 \widetilde{J}_{k,i} \leq   \frac{2 \kappa(\bT) \sqrt{K} \beta  \lambda B}{1-\kappa (\bT)}+\frac{(K-d_{\textup{min}})\: \tau}{1-\kappa (\bT)}   
\end{align}
where \(d_{\textup{min}} \) is the minimum degree over the graph, i.e., minimum number of neighbors any agent over the network possesses, and \( \lambda \triangleq \max \{ |1-\frac{K}{\beta}|, \lambda_2 \} \) where \( \lambda_2 < 1 \) is the mixing rate (second largest modulus eigenvalue) of \( C \) . 
\end{theorem}
\begin{myproof}
 See Appendix~\ref{appendix:risk_theorem}.
\end{myproof}

In Theorem~\ref{th:kl_without_net_assumption}, the first terms in both bounds are equivalent to the bounds obtained in \cite{kayaalp_dist_bayesian}. However, the terms proportional to \( (K-d_{\textup{min}})\tau \) are new, and they arise from the fact that agents do not observe the joint actions and hence only have a local estimate of the transition model. Nevertheless, the bounds get smaller with increasing network connectivity, i.e., as \( \lambda_2 \to 0 \) and \( d_{\textup{min}} \to K \), which shows the benefit of cooperation. In particular, if \( \beta = K \) and the network is fully connected (\( \lambda_2 = 0, d_{\textup{min}} = K \)), then the bounds are equal to $0$. In other words, local beliefs match the centralized belief in this situation. It is important to note that the linear term $(K-d_{\textup{min}})$ represents a worst-case bound that holds true for any strongly connected network topology. For instance, in a scenario where each agent has $N>1$ neighbors, it is straightforward to modify the proof and show that these linear terms will instead be logarithmic, i.e., proportional to $\log K / \log N$.

We use Theorem~\ref{th:kl_without_net_assumption} in the performance analysis of the diffusion policy evaluation. To that regard, we first present the following consequence of Theorem~\ref{th:kl_without_net_assumption}, which provides a bound in terms of disagreement norms. 

\begin{corollary}[{\bf Bounds on disagreement norms}]\label{cor:pinsker}
Theorem~\ref{th:kl_without_net_assumption} implies that, as \( i \to \infty\),
\begin{equation}\label{eq:btv_eq}
       \e \big \|\bmu_i - \bmu_{k,i} \big \| \leq \btv
\end{equation}
and
\begin{equation}\label{eq:btv_kappa_eq}
    \e \big \|\bmeta_i - \bmeta_{k,i} \big \| \leq \wbtv,
\end{equation}
where we introduce the constants
\begin{equation}\label{eq:btv_definition}
    \btv \triangleq  2 \Bigg (1-\exp \Big \{-\frac{2 \sqrt{K} \beta  \lambda B+(K-d_{\textup{min}})\: \tau}{1-\kappa (\bT)} \Big \} \Bigg )^{1/2} 
\end{equation}
and
\begin{equation}\label{eq:wbtv_definition}
    \wbtv \triangleq  2 \Bigg (1-\exp \Big \{-\frac{2 \kappa (\bT) \sqrt{K} \beta  \lambda B+(K-d_{\textup{min}}) \tau}{1-\kappa (\bT)} \Big \} \Bigg )^{1/2}
\end{equation}
\end{corollary}
\begin{myproof}
    See Appendix~\ref{appendix:risk_corollary}.
\end{myproof}
\subsection{Network Disagreement}\label{sec:network_disagreement}
In this section, we study the variation of agent parameters from the network centroid. To that end, let us incorporate the linear approximation \( \widehat{V}(\mu,w) = \phi(\mu)^{\T} w\) into the TD-error expression \eqref{eq:delta_parameter} to obtain the following relation:
\begin{align}
   \bde_{k,i} = \rb_{k,i} + \gamma \phi(\bmeta_{k,i+1})^{\T}\w_{k,i} - \phi(\bmu_{k,i})^{\T}\w_{k,i}.
\end{align}
Since \( \nabla_w \widehat{V} (\mu,w) = \phi (\mu) \) for the linear case, it follows that
\begin{align}
     \bz_{k,i+1} = \Big((1-2\rho\alpha)I-\alpha \bH_{k,i}\Big)\w_{k,i} + \alpha \bD_{k,i},
\end{align}
where
    \begin{equation}\label{eq:hki_definition}
    \bH_{k,i} \triangleq \phi(\bmu_{k,i})\phi(\bmu_{k,i})^{\T}-\gamma \phi(\bmu_{k,i})\phi(\bmeta_{k,i+1})^{\T},
    \end{equation}
and
    \begin{equation}
            \bD_{k,i} \triangleq \rb_{k,i} \phi(\bmu_{k,i}).
    \end{equation}
 To proceed, we introduce the following regularity assumption on the feature vector.

\begin{assumption}[{\bf Feature vector}]\label{assumption:feature}
The feature mapping $\phi(\mu)$ is bounded and Lipschitz continuous in the domain of the \( S\)-dimensional probability simplex. Namely, for any vectors \( \mu_1,\mu_2 \in \mathcal{M}(S) \),
\begin{equation}
    \|\phi(\mu_1) - \phi (\mu_2) \| \leq \lfi \|\mu_1-\mu_2 \|, \quad \| \phi(\mu_1)\| \leq \bfi.
\end{equation}$\qed$
\end{assumption}
\begin{lemma}[{\bf Belief feature difference}]\label{lemma:h_difference}
For each agent \( k \in \mathcal{K}\), the belief feature matrix \( \bH_{k,i} \) in \eqref{eq:hki_definition} has bounded expected difference in relation to the centralized belief feature matrix \( \bH_i^\star \), defined below, i.e.,
\begin{equation}
  \e  \|\bH_{k,i}- \bH_i^\star\| \leq 2\bfi \lfi \btv(1+ \gamma ),
\end{equation}
where
\begin{align}\label{eq:histar_definition}
    \bH_i^\star \triangleq \phi(\bmu_{i})\phi(\bmu_{i})^{\T}-\gamma \phi(\bmu_{i})\phi(\bmeta_{i+1})^{\T} .
\end{align}
\end{lemma}
\begin{myproof}
See Appendix~\ref{sec:appendix_h_difference}.
\end{myproof}
We also assume that all rewards are non-negative and uniformly bounded, i.e., \( 0 \leq \rb_{k,i} \leq \rma \) for each agent \( k \in \mathcal{K} \), and all time instants \( i\). Now, we proceed to study the network disagreement. To that end, we define the network centroid as
\begin{equation}
     \w_{c, i} \triangleq \frac{1}{K} \sum_{k=1}^K  \w_{k, i} ,
\end{equation}
which is an average of the parameters of all agents. The following result shows that the agents cluster around this network centroid after sufficient iterations. 

\begin{theorem}[{\bf Network agreement}]\label{th:network_disagreement} The average distance to the network centroid is bounded for \( \rho > \gamma \bfi \lfi / \sqrt{2} \) after sufficient number of iterations. In particular, if \( \rho \geq 0.75 \gamma \bfi \lfi \), then
\begin{equation}
    \frac{1}{K} \sum_{k=1}^K \e \| \w_{k,i} - \w_{c,i}\| \leq  \frac{\alpha \lambda_2 \epsilon}{(1-\lambda_2)} + O(\alpha^2)
\end{equation}
where \( \epsilon > 0 \) is a constant defined by
\begin{equation}
    \epsilon \triangleq  \rma \bfi \Big (\frac{2\btv(1+ \gamma)}{0.08\gamma}+1 \Big )  .
\end{equation}
\end{theorem}
\begin{myproof}
See Appendix~\ref{sec:appendix_network_disagreement}.
\end{myproof}

Theorem~\ref{th:network_disagreement} states that the parameter estimates by the agents cluster around the network centroid within mean \(\ell_2\)-distance on the order of \( O(\alpha \lambda_2)\) in the limit as $i\rightarrow \infty$. This result confirms that agents can get arbitrarily close to each other by setting the learning rate \( \alpha \) sufficiently small. Besides, dense networks have in general small \( \lambda_2 \), which results in a small disagreement within the network.

\subsection{Performance of Diffusion Policy Evaluation}\label{sec:centralized_disagreement}

We can therefore use the network centroid as a proxy for all agents to show that the disagreement between the fully decentralized strategy of Alg.~\ref{alg:fully_decent} and a baseline strategy that requires a central processor during training is bounded. We start by describing this baseline strategy and explain why it is a more suitable baseline compared to using the fully centralized strategy Alg.~\ref{alg:fully_cent}.

In some applications, even though agents are supposed to work in a decentralized fashion once implemented in the field, they can nevertheless rely on central processing during the training phase in order to learn the best policy. In the literature, this paradigm is referred to as \emph{centralized training for decentralized execution} \cite{foerster2018counterfactual,jorge2016learning}. For our problem, the crucial point is that during training the centralized processor can form beliefs based on all observations, but it should keep in mind that agents will execute their actions based on \emph{local} beliefs once implemented. Therefore, in the baseline strategy, actions and rewards are based on local beliefs as in \eqref{eq:dif_adapt_step}--\eqref{eq:dif_evolve_step}, whereas parameter updates are based on the centralized posterior as in \eqref{eq:centralized_posterior_ma}--\eqref{eq:ck_cent_ma}. Algorithm~\ref{alg:cent_dec_baseline} lists this baseline procedure. Notice that the algorithm consists of both local belief construction (see \eqref{eq:dif_adapt_step_baseline}, \eqref{eq:dif_combine_step_baseline}, and \eqref{eq:dif_evolve_step_baseline}) and centralized belief construction (see \eqref{eq:centralized_posterior_ma_baseline} and \eqref{eq:centralized_posterior_evolve_baseline}). The former is used for action execution \( \ac_{k,i} \sim \pi_k (a_k|\bmu_{k,i}) \), while the latter is used for value function parameter updates in \eqref{eq:cent_pol_learn_del_baseline}--\eqref{eq:cent_pol_learn_w_baseline}.

  \begin{algorithm}[]
 \caption{Centralized evaluation for decentralized execution}
 \begin{algorithmic}[1]
  \item set initial priors $\eta_{k,0}(s)>0,\eta_{0}(s)>0$, for  $\forall s \in\mathcal{S}$ and  $\forall k\in \mathcal{K}$ 
  \item choose $\beta>0$
  \item initialize $w_{0}^\star$
    \While{$i\geq 0$} 
  \State each agent $k$ observes $\bxi_{k,i}$ 
 \FORC{each agent $k\in\mathcal{K}$ and $s \in \mathcal{S}$}
 \begin{align}\label{eq:dif_adapt_step_baseline}
    \bpsi_{k,i} (s) \propto (L_k(\bxi_{k,i} | s))^{\beta}\bmeta_{k,i} (s)  
\end{align}
\begin{equation}\label{eq:dif_combine_step_baseline}
    \bmu_{k,i}(s) \propto \prod_{\ell \in \mathcal{N}_k}  \big ( \bpsi_{\ell,i} (s) \big )^{a_{\ell k}}    
\end{equation}
\ENDFORC
\State to form centralized belief with joint observation \( \bxi_i \triangleq \{\bxi_{k,i}\}_{k=1}^K\), \textbf{adapt}
\begin{equation}\label{eq:centralized_posterior_ma_baseline}
 \bmu_i (s) \propto L(\bxi_i|s) \bmeta_i (s)
\end{equation}
 \For{each agent $k\in\mathcal{K}$}
\State Take action \( \ac_{k,i} \sim \pi_k (a_k|\bmu_{k,i}) \) \label{state:take_actions_baseline}
\State Get reward  \( \rb_{k,i} = \bm{r}_k (\s_i,\ac_{i},\s_{i+1}) \)
\EndFor
\State \textbf{average} the rewards  \( \rb_i^\star = \frac{1}{K} \sum_{k=1}^K \bm{r}_{k,i} \)
\FORB{each agent $k\in\mathcal{K}$} 
 \State Compute \( \bT^\pi_k(s| s^\prime, \ac_{{\cal N}_k,i}) \) using \eqref{eq:approximate_trans_prof}, and
\begin{equation}\label{eq:dif_evolve_step_baseline}
  \bmeta_{k,i+1} (s) = \sum_{s^\prime \in \mathcal{S}} \bT^\pi_k(s| s^\prime, \ac_{{\cal N}_k,i}) \bmu_{k,i}(s^\prime)
\end{equation}
\ENDFORB
\State \textbf{evolve} the centralized belief
\begin{equation}\label{eq:centralized_posterior_evolve_baseline}
\bmeta_{i+1} (s) = \sum_{s^\prime\in\mathcal{S}} \bT(s| s^\prime,\ac_{i}) \bmu_{i}(s^\prime)
\end{equation}
 \State \textbf{update} value function parameter
 \begin{equation}\label{eq:cent_pol_learn_del_baseline}
  \bde_i^\star = \rb_i^\star + \gamma \widehat{V} (\bmeta_{i+1},\w_i^\star) - \widehat{V} (\bmu_i,\w_i^\star)
\end{equation}
\begin{equation}\label{eq:cent_pol_learn_w_baseline}
     \w_{i+1}^\star = (1-2\rho\alpha)\w_i^\star + \alpha \bde_i^\star \nabla_w \widehat{V} (\bmu_i,\w_i^\star)
\end{equation}
 \EndWhile
 \end{algorithmic} \label{alg:cent_dec_baseline}
 \end{algorithm}

In the fully centralized strategy of Alg.~\ref{alg:fully_cent}, the actions by the agents and the subsequent rewards are based on the centralized belief. Therefore, the target value function that Alg.~\ref{alg:fully_cent} aims to learn corresponds to the average cumulative reward obtained under centralized execution. In comparison, the target value functions that Algs. \ref{alg:fully_decent} and \ref{alg:cent_dec_baseline} try to learn are the same and they correspond to the average cumulative reward under decentralized execution. While trying to learn the same parameter $w^\circ$, the baseline strategy can utilize centralized processing, but the diffusion strategy is fully decentralized. Nonetheless, the following result illustrates that the expected disagreement between the baseline strategy and the fully decentralized strategy remains bounded.

\begin{theorem}[{\bf Disagreement with the baseline solution}]\label{th:centralized_disagreement} The expected distance between the baseline strategy and the network centroid is bounded after sufficient iterations for \( \rho > \gamma \bfi \lfi/ \sqrt{2}  \). In particular, if \( \rho \geq 0.75 \gamma \bfi \lfi \), then
\begin{align}\label{eq:disagreement_central_theorem}
    \e \|  \w_{i}^\star - \w_{c,i}\| \leq  \frac{\btv \rma \epsilon^\prime}{0.08 \gamma \bfi \lfi } 
\end{align}
after \(i\geq i_0=  o\left(1/(\alpha \gamma\bfi \lfi )\right) \) iterations, where \( \epsilon^\prime > 0 \) is a constant defined by
\begin{equation}
    \epsilon^\prime \triangleq \frac{2\bfi (1+ \gamma)}{0.08 \gamma}  +   \lfi.
\end{equation}
\end{theorem}
\begin{myproof}
See Appendix~\ref{sec:appendix_centralized_dif}.
\end{myproof}

Theorem~\ref{th:centralized_disagreement} implies that the disagreement between the network centroid, around which agents cluster, and the baseline strategy is on the order of \( \btv \). This means that if the local beliefs are similar to the centralized belief, agents get closer to the baseline parameter. In this regard, from the definition \eqref{eq:btv_definition} of \( \btv \), it can be observed that \( \btv \) gets smaller with increasing network connectivity (i.e., decreasing \( \lambda_2\)), as \( \beta \to K\). In fact, it is equal to zero for fully-connected networks with the choice of \( \beta = K\) and \( c_{\ell k} = 1/K\). Therefore, by changing \( \beta\) and \( c_{\ell k} \), the fully decentralized strategy can match the value function estimates of a centralized training strategy that can gather all observations and actions in a fusion center. In the next section, by means of numerical simulations, we further compare the value function estimate accuracies of all Algorithms \ref{alg:fully_cent}, \ref{alg:fully_decent} and \ref{alg:cent_dec_baseline} by using squared Bellman error (SBE).

\section{SIMULATION RESULTS}\label{sec:simulation_results}

\begin{figure*}
    \centering
       \begin{subfigure}{0.3\textwidth}
            \includegraphics[width=\textwidth]{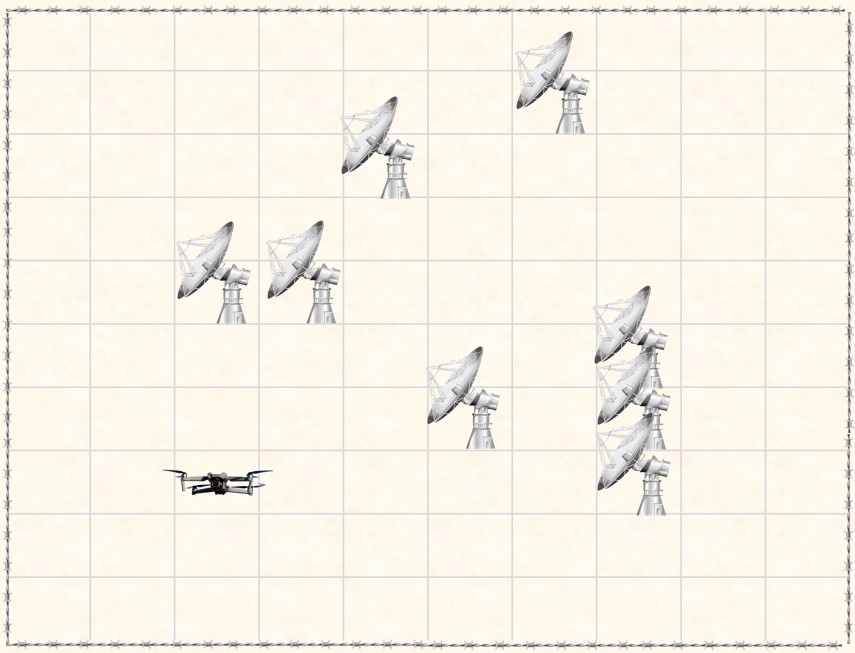}
            \caption{Initial positions at the beginning of an iteration.}
            \label{fig:Fig1} 
        \end{subfigure}
    \begin{subfigure}{0.3\textwidth}
            \includegraphics[width=\textwidth]{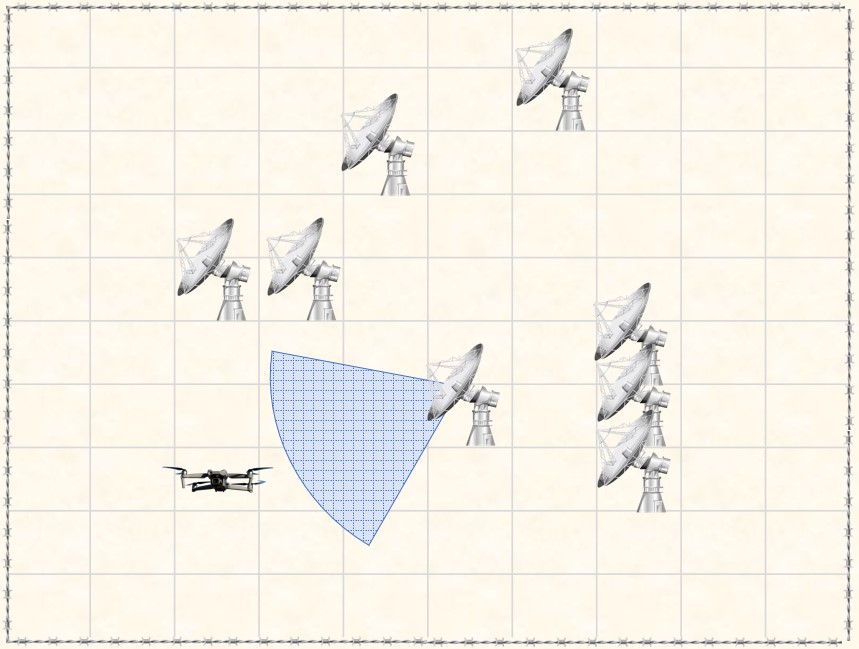}
            \caption{Agents receive noisy observations and incorporate them into their beliefs.}
            \label{fig:Fig2} 
        \end{subfigure}
    \begin{subfigure}{0.3\textwidth}
            \includegraphics[width=\textwidth]{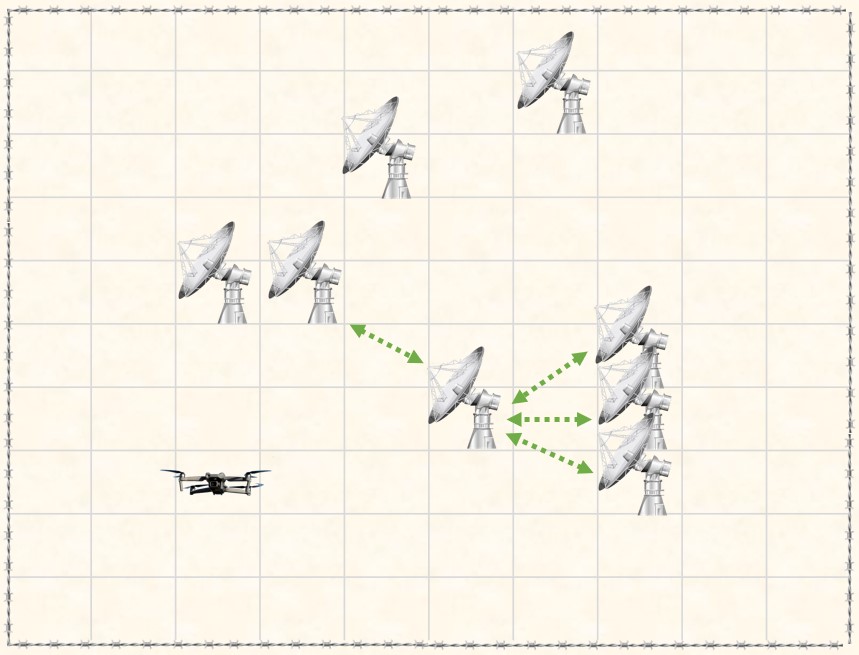}
            \caption{Agents exchange beliefs with their immediate neighbors.}
            \label{fig:Fig3} 
        \end{subfigure}
    \begin{subfigure}[t]{0.3\textwidth}
            \includegraphics[width=\textwidth]{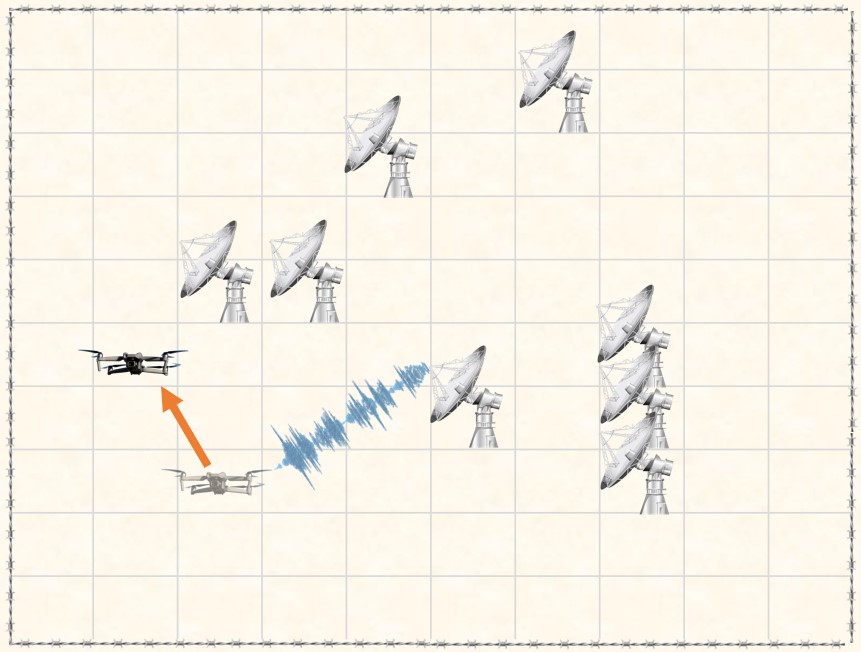}
            \caption{Agents take actions based on the beliefs. The target relocates based on the actions.}
            \label{fig:Fig4} 
        \end{subfigure}
    \begin{subfigure}[t]{0.3\textwidth}
            \includegraphics[width=\textwidth]{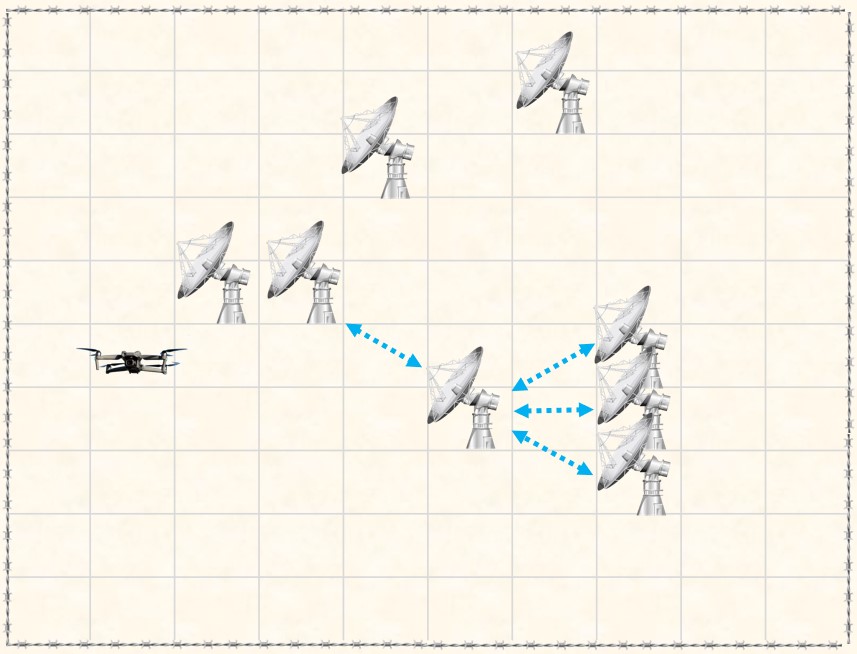}
            \caption{Agents update and exchange value function parameters.}
            \label{fig:Fig5} 
        \end{subfigure}
            \begin{subfigure}[t]{0.3\textwidth}
           \includegraphics[width=\textwidth,height=0.76\textwidth]{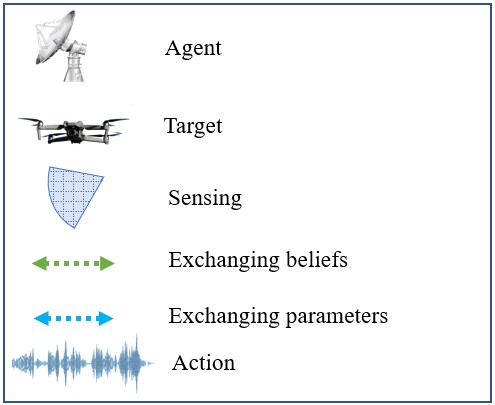} 
            \caption{The legend. Image credit for agent, target, and action: freepik.com}
            \label{fig:Fig6} 
        \end{subfigure}
    \caption{Experimental scenario. For visual purposes, the procedure is shown for only one agent. In fact, all agents execute the same procedure simultaneously.}
    \label{fig:scenario}
\end{figure*}

For numerical simulations, we consider a multi-agent target localization application. The implementation is available online\footnote{\href{https://github.com/asl-epfl/DecPOMDP_Policy_Evaluation_w-Belief_Sharing}{github.com/asl-epfl/DecPOMDP\_Policy\_Evaluation\_w-Belief\_Sharing}}. We use a set of \( K=8 \) agents and a moving target in a \( 10 \times 10 \) two-dimensional grid world environment. The locations of the agents are fixed and their coordinates are randomly assigned at the beginning of the simulation. The goal of the agents is to cooperatively evaluate a given policy for hitting the target. Agents cannot observe the location (i.e., state) of the target accurately, but instead receive noisy observations based on how far they are from the real location of the target. The target is moving according to some pre-defined transition model that takes the actions (i.e., hits) of agents into account. Specifically, the target is trying to evade the hits of agents.

A possible scenario for this setting is a network of sensors and an intruder (e.g., a spy drone) --- see Fig.~\ref{fig:scenario}. The sensors try to localize the intruder based on noisy measurements and belief exchanges. Moreover, in order to disrupt the communication between the intruder and its owner, each sensor sends a narrow sector jamming beam towards its target location estimate. However, the intruder is capable of detecting energy abnormalities and determines its next location by favoring distant locations from the jamming signals. We now describe the setting in more detail. \\

\begin{figure*}
    \centering
       \begin{subfigure}[t]{0.25\textwidth}
    \includegraphics[width=\textwidth, height=1.2\textwidth]{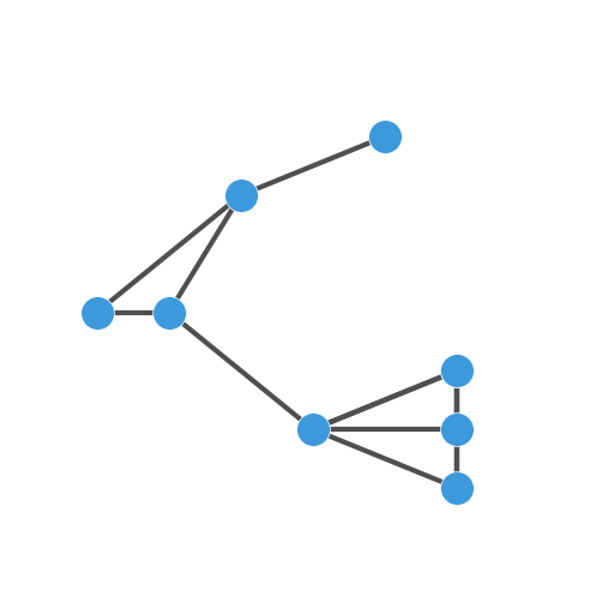}
    \caption{Communication topology.}
    \label{fig:exp_network}
        \end{subfigure}
    \begin{subfigure}[t]{0.33\textwidth}
    \includegraphics[width=\textwidth]{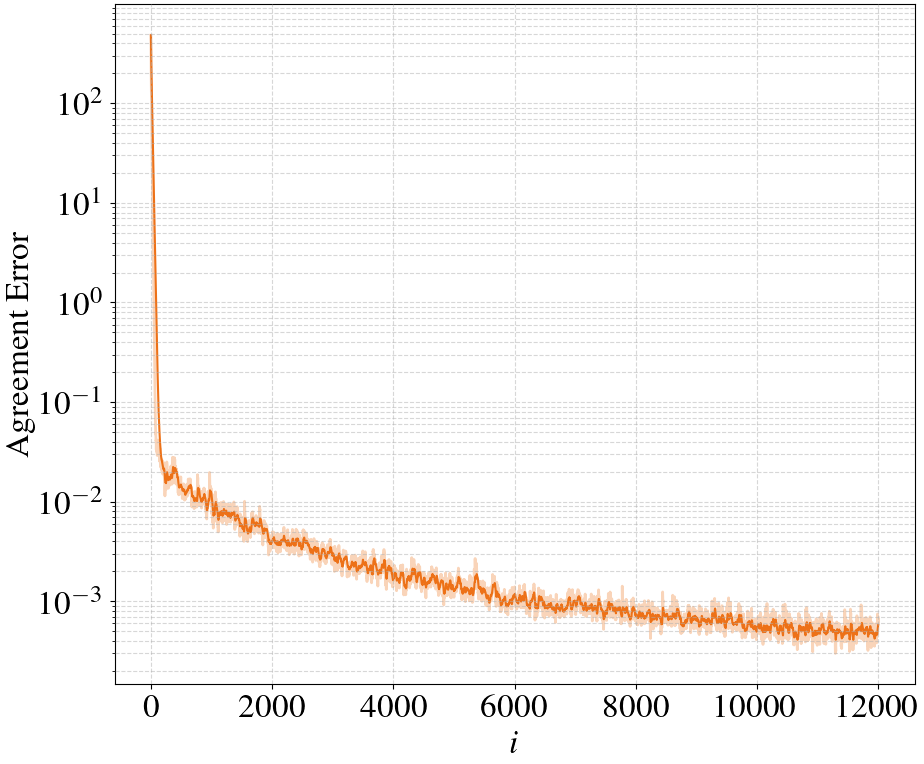}
    \caption{Agreement error over time.}
    \label{fig:agreement}
        \end{subfigure}
    \begin{subfigure}[t]{0.33\textwidth}
\includegraphics[width=\textwidth]{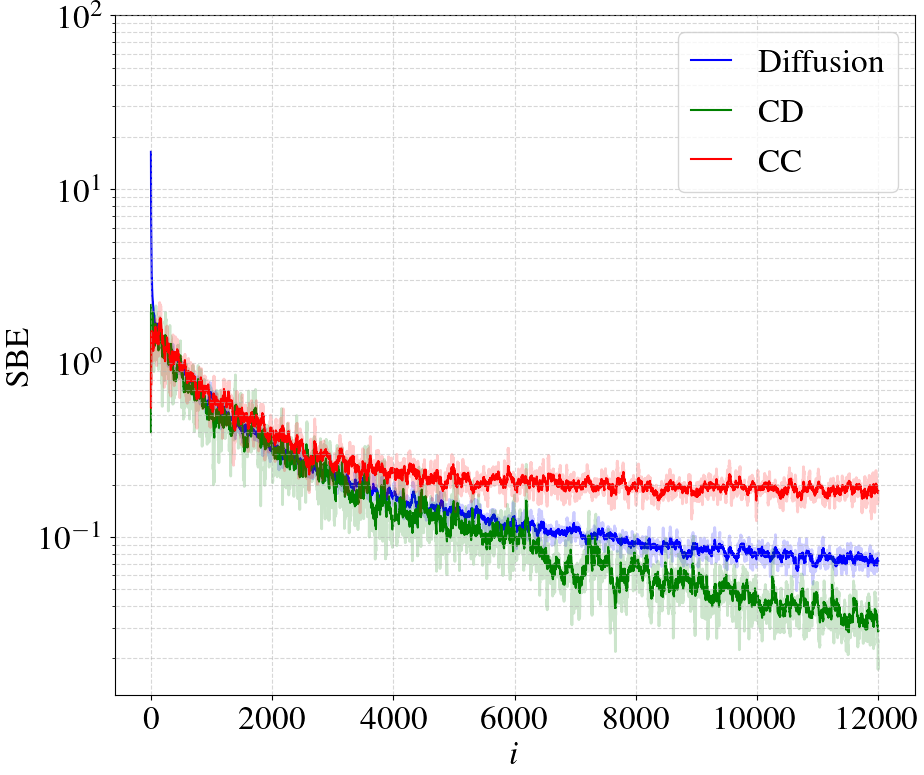}
  \caption{SBE over time (running window of size 20) for CC (Alg.~\ref{alg:fully_cent}), Diffusion (Alg.~\ref{alg:fully_decent}), CD (Alg.~\ref{alg:cent_dec_baseline}).}
  \label{fig:sbe_two} 
        \end{subfigure}
        \caption{}
    \label{fig:plots}
\end{figure*}

    \noindent \textbf{Combination matrix:} The entries of the combination matrix are set such that they are inversely proportional to the \( \ell_1\)-distance between the agents. That is to say, the further the agents are from each other, the smaller the value of the weight that is assigned to the edge connecting them. Weights smaller than some threshold are set to 0, which implies that agents that are too far from each other do not need to communicate. The resulting communication topology graph is illustrated in Fig.~\ref{fig:exp_network}.

    \noindent \textbf{Transition model:} The target is moving between cells in a grid (i.e., states) randomly. The probability of a cell being the next location of the target depends on the current location of the target and the location of the agents' hits. Namely, each state in the grid is assigned a score based on its \(\ell_1\)-distance to the current location of the target and to the average location of the agents' hits --- see Table~\ref{Table2}. For example, observe from Table~\ref{Table2} that the cells that are in the proximity of the target's current location and also far away from the agents' strikes are given the highest score. These scores are normalized to yield a probabilistic transition kernel.

\renewcommand{\arraystretch}{1.8}
\newcolumntype{B}{ >{\centering\arraybackslash} m{2.6cm} }
\newcolumntype{C}{ >{\centering\arraybackslash} m{1.1cm} }
\newcolumntype{D}{ >{\centering\arraybackslash} m{0.9cm} }
\renewcommand\multirowsetup{\centering}
\begin{table}
\begin{center}
\begin{tabular}{|B|C|D|D|}
\cline{3-4}
\multicolumn{2}{c|}{} & \multicolumn{2}{c|}{\textbf{initial position}} \\
\cline{2-4}
\multicolumn{1}{c|}{} & $\bm{\ell}_\textbf{1}$\textbf{-distance} & $\leq 4$ & $> 4$ \\
\hline
\multirow{2}*{\begin{tabular}{l}\textbf{average location} \\
                                      \textbf{of agents' hits} \\
                \end{tabular} }  & {$< 4$}  & 10 & 5 \\
\cline{2-4}
& $\geq 4$ & 100 & 50 \\
\hline
\end{tabular}
\end{center}
\renewcommand{\arraystretch}{1}
\caption{The table of scores used in the transition model. Each candidate state for next state (location) of the target gets a score based on the initial position of the target and the average action of agents.}
\label{Table2}
\end{table}

     \noindent \textbf{Likelihood function:} Agents cannot observe where the target is. They can only receive noisy observations. Each agent gets a more accurate observation of the target's position if the target is in close proximity to the agent. Otherwise, the larger the distance between the agent and the target, the higher the noise level. Depending on how close the target is to the agent, and in order to construct the likelihood function, we first assign scores to each cell in the grid that reflect how probable it is to find the target in that cell --- see Table~\ref{Table3}. Following that, the scores are normalized in order to yield a distribution function. For instance, if the target lies at an \( \ell_1\)-distance that is less than 3 grid squares from the location of the agent, the actual position of the target gets a likelihood score of 400, cells within an \( \ell_1\) distance of 2 grid squares from the agent get a likelihood score of 200, and cells within an \( \ell_1\) distance of 4 grid squares from the agents get a likelihood score of 30. \\

\newcolumntype{E}{ >{\centering\arraybackslash} m{0.45cm} }
\newcolumntype{F}{ >{\centering\arraybackslash} m{1.3cm} }
\newcolumntype{G}{ >{\centering\arraybackslash} m{0.9cm} }
\newcolumntype{H}{ >{\centering\arraybackslash} m{1.08cm} }
\renewcommand{\arraystretch}{1.7}
\begin{table}
\begin{center} 
\begin{tabular}{|F|H|E|E|E|E|E|E|}
\cline{3-8}
\multicolumn{2}{c|}{} & \multicolumn{6}{c|}{\textbf{real location of target}} \\
\cline{2-8}
\multicolumn{1}{c|}{} & $\bm{\ell}_\textbf{1}$\textbf{-distance} & $= 0$ & $< 3$ & $< 5$ & $<7$ & $<9$ & $\geq 9$ \\
\hline
\multirow{3}*{\begin{tabular}{l}\textbf{location}\\ 
                                      \textbf{of}\\ 
                                      \textbf{agent}\\ 
                \end{tabular} }   & \rule{0pt}{0ex}{$< 3$} & \rule{0pt}{0ex}{400}  & \rule{0pt}{0ex}{200}  & \rule{0pt}{0ex}{30} & \rule{0pt}{0ex}{1} & \rule{0pt}{0ex}{1} & \rule{0pt}{0ex}{1} \\
\cline{2-8}
& \raisebox{0ex}{$\leq 6$} & \raisebox{0ex}{200} & \raisebox{0ex}{180} & \raisebox{0ex}{100} & \raisebox{0ex}{1} & \raisebox{0ex}{1}  & \raisebox{0ex}{1}\\
\cline{2-8}
& \raisebox{0ex}{$> 6$} & \raisebox{0ex}{25} & \raisebox{0ex}{25} & \raisebox{0ex}{25} & \raisebox{0ex}{25} & \raisebox{0ex}{4} & \raisebox{0ex}{1} \\
\hline
\end{tabular} 
\end{center}
\renewcommand{\arraystretch}{1}
\caption{The table of scores used in the likelihood function model. Each state, when observed, gets a score that determines the likelihood of the presence of the target within the state, based on the position of the target and the average action of agents.}
\label{Table3}
\end{table}
\renewcommand{\arraystretch}{1}

    \noindent \textbf{Reward function:} The reward function in the environment is such that an agent receives a reward of 1 if the agent is able to  hit the position of the target. The agent also receives a reward of 0.2 if the \( \ell_1\)-distance between the predicted location and the actual location of the target is less than 3 grid units. Otherwise, it gets 0 reward. Agents do not know the reward model, and use the instantaneous rewards instead.
    
    \noindent \textbf{Policy:} We fix the policy that the agents evaluate as the maximum a-posteriori policy. Namely, agents detect (hit) a location if it corresponds to the maximum entry in their belief vector.

We use the belief vectors as the features directly, i.e., \( \phi\) is an identity transformation. We set \( \alpha = 0.1 \), \( \rho = 0.0001 \), and \(\beta =K=8\), and average over 3 different realizations for all cases.  In Fig.~\ref{fig:agreement}, the average mean-square distance to the network centroid, i.e.,
\begin{equation}
       \text{Agreement error} \triangleq \frac{1}{K} \sum_{k=1}^K \e \| \w_{k,i} - \w_{c,i}\|^2 ,
\end{equation}
is plotted over time for the fully decentralized strategy. Confirming Theorem~\ref{th:network_disagreement}, it can be seen that agreement error rapidly decreases and converges to a small value.

In Fig. \ref{fig:sbe_two}, we plot the evolution of the average squared Bellman error (SBE) in the log domain, where the SBE expression is given by:
\begin{equation}
   \text{SBE} \triangleq \frac{1}{K} \sum_{k=1}^K \bde_{k,i}^2,
\end{equation}
    and similarly for the centralized cases. It measures the network average of instantaneous TD-errors. It can be seen that all approaches converge, and in particular, diffusion strategy (Alg.~\ref{alg:fully_decent}) yields a comparable performance with CD (Alg.~\ref{alg:cent_dec_baseline}). This observation is in line with Theorem~\ref{th:centralized_disagreement}, which states that the disagreement between the fully decentralized strategy and the baseline centralized training for decentralized execution strategy is bounded. Notice also that CC (Alg.~\ref{alg:fully_cent}) results in a higher SBE compared to the diffusion and CD, despite being a fully centralized strategy. This is because, CC evaluates a different policy, namely, the centralized execution policy. Therefore, as argued in Sec.~\ref{sec:theoretical_results}-\ref{sec:centralized_disagreement}, the SBE of CC is not a suitable baseline for the diffusion strategy.

\section{CONCLUDING REMARKS}

 In this paper, we proposed a policy evaluation algorithm for Dec-POMDPs over networks. We carried out a rigorous analysis that established: $(i)$ the beliefs formed with local information and interactions have a bounded disagreement with the global posterior distribution, $(ii)$ agents' value function parameters cluster around the network centroid, and $(iii)$ the decentralized training can match the performance of the centralized training with appropriate parameters and increasing network connectivity. 

There are two limitations of the current work that can be addressed in future work. First, we assume that agents know the local likelihood and transition models accurately. One possible question is if agents have approximation errors for the models, how would these affect the analytical results. Second, an implication of Theorem~\ref{th:centralized_disagreement} is that there is necessity for regularization \((\rho > 0)\). We leave the question of whether one can get bounds that does not require this, possibly with more assumptions on the model, to future work.




\section*{APPENDIX}\label{sec:appendices}

\allowdisplaybreaks
\subsection{Proof of Theorem~\ref{th:kl_without_net_assumption}}\label{appendix:risk_theorem}
We can rewrite the risk function as
\begin{align}
J_{k,i}&=\e_{\f_i} \dkl(\bmu_i || \bmu_{k,i}) \notag\\
&= \e_{\f_i} \Big [\sum_{s \in \mathcal{S}} \bmu_i (s) \log \frac{\bmu_i (s)}{\bmu_{k,i} (s)} \Big] \notag \\
&\stackrel{(a)}{=} \e_{\f_i} \Big [\sum_{s \in \mathcal{S}} \mathbb{P} (\s_i = s | \bmf_i) \log \frac{\bmu_i (s)}{\bmu_{k,i} (s)} \Big] \notag \\
&\stackrel{(b)}{=}\e_{\f_i} \Big [ \e_{s_i | \f_i} \Big (\log \frac{\bmu_i (\s_i)}{\bmu_{k,i} (\s_i)} \Big ) \Big ] \notag \\
&=\e_{\f_i,s_i}\Big [\log \frac{\bmu_i (\s_i)}{\bmu_{k,i} (\s_i)} \Big ],
\end{align}
 where \( (a) \) follows from definition \eqref{eq:true_posterior}, \( (b) \) follows from the definition of conditional expectation with respect to \( \s_i \) given \( \bmf_i\). Merging the diffusion adaptation step \eqref{eq:dif_adapt_step} and the combination step \eqref{eq:dif_combine_step} together yields the following form:
 \begin{equation}
    \bmu_{k,i} (s) \propto \prod_{\ell \in \mathcal{N}_k} (L_\ell(\bxi_{\ell,i} | s))^{\beta c_{\ell k}}(\bmeta_{\ell,i} (s))^{c_{\ell k}} ,
\end{equation}
which, combined with the update equation \eqref{eq:centralized_posterior_ma} for the centralized solution, results in:
\begin{align}\label{eq:log_ratio_expansion}
    \log &\frac{\bmu_i (s)}{\bmu_{k,i} (s)}  = \sum_{\ell \in \mathcal{N}_k} c_{\ell k} \Big ( \log \frac{L(\bxi_i|s)}{(L_\ell(\bxi_{\ell,i}|s))^{\beta}}+\log \frac{\bmeta_i(s)}{\bmeta_{\ell,i}(s)} \Big ) \notag \\
    &  +  \log \sum_{s^\prime \in \mathcal{S}} \!\! \Big( \!\!\prod_{\ell \in \mathcal{N}_k}\! (L_\ell(\bxi_{\ell,i} | s^\prime))^{\beta c_{\ell k}} \!\prod_{\ell \in \mathcal{N}_k} \! (\bmeta_{\ell,i} (s^\prime))^{c_{\ell k }} \Big )  \notag\\
&\qquad  -  \log  \m_i(\bxi_i)  .
\end{align}
Here, we have introduced the marginal distribution of new observation given the past observations and actions:
\begin{align}
\m_i(\xi_i) \triangleq \mathbb{P} (\bxi_i=\xi_i | \bmf_{i-1}^a ) &= \sum_{s \in \mathcal{S}} \mathbb{P} (\bxi_i=\xi_i, \s_i =s | \bmf_{i-1}^a ) \notag \\ &= \sum_{s \in \mathcal{S}} L(\xi_i|s)   \mathbb{P} (\s_i =s | \bmf_{i-1}^a ) \notag \\ &= \sum_{s \in \mathcal{S}} L(\xi_i|s) \bmeta_i(s).
\end{align}

First, observe that the expectation of the log-likelihood ratio terms in \eqref{eq:log_ratio_expansion} satisfies:
\begingroup
\allowdisplaybreaks
\begin{align}
\sum_{\ell \in \mathcal{N}_k} c_{\ell k} &\e_{\xi_i,s_i} \Big [\log \frac{L(\bxi_i|\s_i)}{(L_\ell(\bxi_{\ell,i}|\s_i))^{\beta}} \Big ]\notag \\
& \stackrel{(a)}{=} \e_{\xi_i,s_i} \Big[ \sum_{\ell=1}^K \log L_\ell(\bxi_{\ell,i} | \s_{i}) \Big ] \notag \\ &\qquad- \sum_{\ell \in \mathcal{N}_k} c_{\ell k} \e_{\xi_{\ell,i},s_i} \Big[ \beta  \log L_\ell(\bxi_{\ell,i} | \s_{i}) \Big]  \notag\\
& =  \e_{\xi_i,s_i} \Big[ \sum_{\ell=1}^K (1  - \beta c_{\ell k}) \log L_\ell(\bxi_{\ell,i} | \s_{i})  \Big] \label{eq:first_4_term}
\end{align}
\endgroup
where in \( (a) \) we used the spatial independency of the observations. Second, the expectation of the time-adjusted terms in \eqref{eq:log_ratio_expansion} can be rewritten as:
\begin{align}\label{eq:time_adj_expansion}
    &\sum_{\ell \in \mathcal{N}_k} c_{\ell k} \e_{\f_i,s_i} \Big[  \log \frac{\bmeta_i(\s_i)}{\bmeta_{\ell,i}(\s_i)} \Big ] \notag \\& \stackrel{(a)}{=}  \sum_{\ell \in \mathcal{N}_k}  c_{\ell k} \e_{\f_i,s_i} \Big[  \log \frac{\bmeta_i(\s_i)}{\bwmeta_{\ell,i}(\s_i)}+ \log \frac{\bwmeta_{\ell,i}(\s_i)}{\bmeta_{\ell,i}(\s_i)} \Big ] \notag \\
    &  =\!\! \sum_{\ell \in \mathcal{N}_k} \!\! c_{\ell k} \e_{\f_{i-1}^a, s_i} \Big[ \e_{\xi_i|\f_{i-1}^a, s_i} \Big (  \!\log \frac{\bmeta_i(\s_i)}{\bwmeta_{\ell,i}(\s_i)}+ \log \frac{\bwmeta_{\ell,i}(\s_i)}{\bmeta_{\ell,i}(\s_i)}  \Big ) \!\Big ] \notag \\
    &  \stackrel{(b)}{=}  \sum_{\ell \in \mathcal{N}_k}  c_{\ell k} \e_{\f_{i-1}^a,s_i} \Big[  \log \frac{\bmeta_i(\s_i)}{\bwmeta_{\ell,i}(\s_i)}+ \log \frac{\bwmeta_{\ell,i}(\s_i)}{\bmeta_{\ell,i}(\s_i)} \Big ] 
\end{align}
where in \( (a) \) we define the agent-specific distribution:
\begin{equation}
    \bwmeta_{\ell,i}(s) \triangleq \sum_{s^\prime \in \mathcal{S}}\bT(s| s^\prime,\ac_{i-1}) \bmu_{\ell,i-1}(s^\prime),
\end{equation}
and \( (b) \) follows from the fact that the arguments are deterministic given the current state and the history of actions and observations. The first term of \eqref{eq:time_adj_expansion} can be written as a KL-divergence because of the following:
\begingroup
\allowdisplaybreaks
\begin{align}
 &\sum_{\ell \in \mathcal{N}_k} c_{\ell k} \e_{\f_{i-1}^a,s_i} \Big[  \log \frac{\bmeta_i(\s_i)}{\bwmeta_{\ell,i}(\s_i)} \Big ] \notag\\
&\qquad = \sum_{\ell \in \mathcal{N}_k} c_{\ell k} \e_{\f_{i-1}^a} \Big[ \e_{s_i | \f_{i-1}^a}  \Big(  \log \frac{\bmeta_i(\s_i)}{\bwmeta_{\ell,i}(\s_i)} \Big ) \Big ] \notag\\
&\qquad = \sum_{\ell \in \mathcal{N}_k} c_{\ell k} \e_{\f_{i-1}^a} \Big[\sum_{s \in \mathcal{S}} \mathbb{P} (\s_i = s | \bmf_{i-1}^a) \log \frac{\bmeta_i(s)}{\bwmeta_{\ell,i}(s)} \Big ] \notag\\
&\qquad \stackrel{\eqref{eq:ck_cent}}{=} \sum_{\ell \in \mathcal{N}_k} c_{\ell k} \e_{\f_{i-1}^a} \Big[\sum_{s \in \mathcal{S}} \bmeta_i(s) \log \frac{\bmeta_i(s)}{\bwmeta_{\ell,i}(s)} \Big ] \notag\\
&\qquad = \sum_{\ell \in \mathcal{N}_k} c_{\ell k} \e_{\f_{i-1}^a} \Big [\dkl (\bmeta_i || \bwmeta_{\ell,i} ) \Big ] .
\end{align}
\endgroup
This expected KL-divergence can be bounded by using the strong-data processing inequality \cite{polyanskiy_2017}:
\begin{align}
    \sum_{\ell \in \mathcal{N}_k} &c_{\ell k} \e_{\f_{i-1}^a} \Big [\dkl (\bmeta_i || \bwmeta_{\ell,i} ) \Big ] \notag \\ & \leq \sum_{\ell \in \mathcal{N}_k} c_{\ell k} \kappa (\bT) \underbrace{\e_{\f_{i-1}} \Big [\dkl (\bmu_{i-1} || \bmu_{\ell,i-1} ) \Big ]}_{J_{\ell,i-1}} \label{eq:second_4_term}.
\end{align}
The second term of \eqref{eq:time_adj_expansion} arises due to transition model disagreement with the centralized belief. To bound it, we first introduce the LogSumExp function \( f \) with vector arguments \( \nu \in \mathbb{R}^S\):
\begin{align}\label{eq:f_logsumexp_def}
    f(\nu) \triangleq \log \sum_{s \in \mathcal{S}} \exp \{\nu (s) \}.
\end{align}
Its gradient is given by
\begin{equation}\label{eq:logsumexp_grad_def_1}
    \nabla_{\nu} f ( \nu) \triangleq \mcl \Big \{ \frac{\partial f(\nu)}{\partial  \nu (s)} \Big\}_{s \in \mathcal{S}}=\mcl \Big \{ \frac{\exp \{\nu(s)\}}{\sum_{s^\prime} \exp \{\nu(s^\prime)\}} \Big \}_{s \in \mathcal{S}}.
\end{equation}
Observe that if we define the vectors
\begin{equation}\label{eq:tilde_nu_def}
    \widetilde{\bnu}_{\ell,i} \triangleq \mcl \Big   \{ \log \big (\bT(\s_i| s,\ac_{i-1}) \bmu_{\ell,i-1}(s)\big ) \Big \}_{s \in \mathcal{S}}
\end{equation}
and
\begin{equation}\label{eq:nu_def}
    \bnu_{\ell,i} \triangleq \mcl\Big \{ \log \big (\bT^\pi_\ell(\s_i| s, \ac_{{\cal N}_\ell,i-1}) \bmu_{\ell,i-1}(s)\big ) \Big \}_{s \in \mathcal{S}} ,
\end{equation}
then, we can rewrite the second expression of \eqref{eq:time_adj_expansion} as follows:
\begin{align}
    &\sum_{\ell \in \mathcal{N}_k} c_{\ell k} \e_{\f_{i-1}^a,s_i} \Big[  \log \frac{\bwmeta_{\ell,i}(\s_i)}{\bmeta_{\ell,i}(\s_i)} \Big ] \notag \\&\qquad = \sum_{\ell \in \mathcal{N}_k} c_{\ell k} \e_{\f_{i-1}^a,s_i} \Big[ f(\widetilde{\bnu}_{\ell,i})-f(\bnu_{\ell,i}) \Big ]. \label{eq:trans_disagree_f}
\end{align}
Applying mean value theorem to this difference yields
\begin{align}
   &\e_{\f_{i-1}^a,s_i} \Big[ f(\widetilde{\bnu}_{\ell,i})-f(\bnu_{\ell,i}) \Big ]\notag \\ &=  \e_{\f_{i-1}^a,s_i} \bigg [ (\nabla_{\nu} f ( \overline{\bnu}_{\ell,i}))^{\T}   \cdot (\widetilde{\bnu}_{\ell,i} - \bnu_{\ell,i})\bigg ] \notag \\
&\quad \stackrel{\eqref{eq:logsumexp_grad_def_1}}{=}\e_{\f_{i-1}^a,s_i} \bigg [ \mcl \Big \{ \frac{\exp \{ \overline{\bnu}_{\ell,i}(s) \}}{\sum_{s^\prime} \exp \{ \overline{\bnu}_{\ell,i}(s^\prime) \} } \Big \}^{\T}_{s \in \mathcal{S}} \cdot (\widetilde{\bnu}_{\ell,i} - \bnu_{\ell,i})  \bigg ] \notag \\&\stackrel{\eqref{eq:tilde_nu_def},\eqref{eq:nu_def}}{=}\e_{\f_{i-1}^a,s_i} \bigg [ \mcl \Big \{ \frac{\exp \{ \overline{\bnu}_{\ell,i}(s) \}}{\sum_{s^\prime} \exp \{ \overline{\bnu}_{\ell,i}(s^\prime) \} } \Big \}^{\T}_{s \in \mathcal{S}} \notag \\ &\qquad \qquad \qquad \cdot \mcl \Big \{ \log \frac{\bT(\s_i| s,\ac_{i-1})}{\bT^\pi_\ell(\s_i| s, \ac_{{\cal N}_\ell,i-1})} \Big\}_{s \in \mathcal{S}} \bigg ] \label{eq:first_mean_val_1} 
\end{align}
for some  \( \overline{\bnu}_{\ell,i} \) between \( \widetilde{\bnu}_{\ell,i}  \) and \(  \bnu_{\ell,i} \). The term in \eqref{eq:first_mean_val_1} is bounded as follows:
\begin{align}
&\Bigg | \e_{\f_{i-1}^a,s_i} \bigg [ \mcl \Big \{ \frac{\exp \{ \overline{\bnu}_{\ell,i}(s) \}}{\sum_{s^\prime} \exp \{ \overline{\bnu}_{\ell,i}(s^\prime) \} } \Big \}^{\T}_{s \in \mathcal{S}} \notag \\ &\qquad \qquad \qquad \cdot \mcl \Big \{ \log \frac{\bT(\s_i| s,\ac_{i-1})}{\bT^\pi_\ell(\s_i| s, \ac_{{\cal N}_\ell,i-1})} \Big\}_{s \in \mathcal{S}} \bigg ] \Bigg |  \notag \\
 &\stackrel{(a)}{\leq}  \e_{\f_{i-1}^a,s_i} \Bigg |  \mcl \Big \{ \frac{\exp \{ \overline{\bnu}_{\ell,i}(s) \}}{\sum_{s^\prime} \exp \{ \overline{\bnu}_{\ell,i}(s^\prime) \} } \Big \}^{\T}_{s \in \mathcal{S}} \notag \\ &\qquad \qquad \qquad \cdot \mcl \Big \{ \log \frac{\bT(\s_i| s,\ac_{i-1})}{\bT^\pi_\ell(\s_i| s, \ac_{{\cal N}_\ell,i-1})} \Big\}_{s \in \mathcal{S}}  \Bigg |  \notag \\
  &\stackrel{(b)}{\leq}  \e_{\f_{i-1}^a,s_i} \Bigg [\Big \|  \mcl \Big \{ \frac{\exp \{ \overline{\bnu}_{\ell,i}(s) \}}{\sum_{s^\prime} \exp \{ \overline{\bnu}_{\ell,i}(s^\prime) \} } \Big \}_{s \in \mathcal{S}} \Big \|_1 \notag \\ &\qquad \qquad \qquad \cdot \Big \|  \mcl \Big \{ \log \frac{\bT(\s_i| s,\ac_{i-1})}{\bT^\pi_\ell(\s_i| s, \ac_{{\cal N}_\ell,i-1})} \Big\}_{s \in \mathcal{S}}  \Big \|_{\infty} \Bigg ] \notag \\
 &\stackrel{(c)}{=} \e_{s_i,a_{i-1}} \Bigg \| \mcl \Big \{ \log \frac{\bT(\s_i| s,\ac_{i-1})}{\bT^\pi_\ell(\s_i| s, \ac_{{\cal N}_\ell,i-1})} \Big\}_{s \in \mathcal{S}}  \Bigg \|_{\infty} \label{eq:log_trans_difference}
 \end{align}
where \( (a) \) follows from the Jensen's inequality, \( (b) \) follows from the Hölder's inequality, and \( (c) \) follows from the fact that
\begin{align}
    \Bigg \| \mcl \Big \{ \frac{\exp \{  \overline{\bnu}_{\ell,i}(s) \}}{\sum_{s^\prime \in \mathcal{S}} \exp \{\overline{\bnu}_{\ell,i}(s^\prime) \} } \Big \} \Bigg \|_1 = 1.
\end{align}

Furthermore, due to Assumption~\ref{assumption:trans_disagreement} and to the fact that the number of maximum hops outside $\mathcal{N}_k$ is $(K- |\mathcal{N}_k|)$, we have
\begin{align}
    \Bigg | \log \frac{\bT(\s_i| s,\ac_{i-1})}{\bT^\pi_k(\s_i| s, \ac_{{\cal N}_k,i-1})} \Bigg | &\leq (K-|\mathcal{N}_k|)\: \tau \notag \\
    &\leq (K-d_{\textup{min}})\: \tau. \label{eq:dmin_bound}
\end{align}

If we combine \eqref{eq:second_4_term}, \eqref{eq:trans_disagree_f}, and \eqref{eq:dmin_bound}, the expectation of the time-adjusted terms in \eqref{eq:log_ratio_expansion} can be bounded as:
\begin{align}\label{eq:prior_wrt_risk}
    &\sum_{\ell \in \mathcal{N}_k} c_{\ell k} \e_{\f_{i-1}^a,s_i} \Big[  \log \frac{\bmeta_i(\s_i)}{\bmeta_{\ell,i}(\s_i)} \Big ] \notag \\& \qquad \quad \leq (K-d_{\textup{min}})\: \tau + \sum_{\ell \in \mathcal{N}_k} c_{\ell k} \kappa (\bT) J_{\ell,i-1}
\end{align}

Next, we bound the expectation of the remaining normalization terms in \eqref{eq:log_ratio_expansion}, which follows similar steps to what was done in \cite{kayaalp_dist_bayesian}:
\begingroup
\allowdisplaybreaks
\begin{align}
&\e_{\f_i} \Bigg [\log \sum_{s^\prime \in \mathcal{S}} \!\! \Big( \!\!\prod_{\ell \in \mathcal{N}_k}\! (L_\ell(\bxi_{\ell,i} | s^\prime))^{\beta c_{\ell k}} \!\prod_{\ell \in \mathcal{N}_k} \! (\bmeta_{\ell,i} (s^\prime))^{c_{\ell k }} \Big ) \Bigg ] \notag\\
&\quad  -  \e_{\f_i} \Big [\log  \m_i(\bxi_i) \Big ] \notag \\
&\qquad \stackrel{(a)}{\leq} \e_{\f_i} \Bigg [\log \sum_{s^\prime \in \mathcal{S}} \!\! \Big( \!\!\prod_{\ell \in \mathcal{N}_k}\! (L_\ell(\bxi_{\ell,i} | s^\prime))^{\beta c_{\ell k}} \!\sum_{\ell \in \mathcal{N}_k} \! c_{\ell k } \bmeta_{\ell,i} (s^\prime) \Big ) \Bigg ] \notag\\
&\qquad  -  \e_{\f_i} \Big [\log  \m_i(\bxi_i) \Big ] \notag \\
&\qquad = \e_{\f_i} \Bigg [\log \sum_{s^\prime \in \mathcal{S}} \!\! \Big( \!\!\prod_{\ell \in \mathcal{N}_k}\! (L_\ell(\bxi_{\ell,i} | s^\prime))^{\beta c_{\ell k}} \!\sum_{\ell \in \mathcal{N}_k} \! c_{\ell k } \bmeta_{\ell,i} (s^\prime) \Big ) \Bigg ] \notag\\
&\qquad - \e_{\f_i} \Bigg [\log \sum_{s^\prime \in \mathcal{S}} \!\! \Big( \prod_{\ell =1}^K\! L_\ell(\bxi_{\ell,i} | s^\prime) \!\sum_{\ell \in \mathcal{N}_k} \! c_{\ell k } \bmeta_{\ell,i} (s^\prime) \Big ) \Bigg ] \notag\\
&\qquad + \e_{\f_i} \Bigg [\log \sum_{s^\prime \in \mathcal{S}} \!\! \Big( \prod_{\ell =1}^K \! L_\ell(\bxi_{\ell,i} | s^\prime) \!\sum_{\ell \in \mathcal{N}_k} \! c_{\ell k } \bmeta_{\ell,i} (s^\prime) \Big ) \Bigg ] \notag\\
&\qquad  -  \e_{\f_i} \Big [\log  \m_i(\bxi_i) \Big ] \notag \\
&\qquad \stackrel{(b)}{\leq} \e_{\f_i} \Bigg [\log \sum_{s^\prime \in \mathcal{S}} \!\! \Big( \prod_{\ell=1}^K\! (L_\ell(\bxi_{\ell,i} | s^\prime))^{\beta c_{\ell k}} \!\sum_{\ell \in \mathcal{N}_k} \! c_{\ell k } \bmeta_{\ell,i} (s^\prime) \Big ) \Bigg ] \notag\\
&\qquad - \e_{\f_i} \Bigg [\log \sum_{s^\prime \in \mathcal{S}} \!\! \Big( \prod_{\ell =1}^K\! L_\ell(\bxi_{\ell,i} | s^\prime) \!\sum_{\ell \in \mathcal{N}_k} \! c_{\ell k } \bmeta_{\ell,i} (s^\prime) \Big ) \Bigg ] \label{eq:nus_before_defined}
\end{align}
\endgroup
where \( (a) \) follows from the arithmetic-geometric mean inequality, \( (b) \) follows from:
\begin{align}
 &- \e_{\f_i} \Bigg [\log \frac{ \m_i(\bxi_i)}{ \sum_{s^\prime \in \mathcal{S}} \!\! \Big( \prod_{\ell =1}^K \! L_\ell(\bxi_{\ell,i} | s^\prime) \!\sum_{\ell \in \mathcal{N}_k} \! c_{\ell k } \bmeta_{\ell,i} (s^\prime) \Big ) }  \Bigg ] \notag\\
 &\qquad \qquad=-\e_{\f_{i-1}^a} \e_{\xi_i| \f_{i-1}^a} \Bigg [\log \frac{ \m_i(\bxi_i)}{\m_i^\dagger(\bxi_i)}  \Bigg ] \notag\\
 &\qquad \qquad = - \e_{\f_{i-1}^a} \dkl(\m_i(\bxi_i) || \m_i^\dagger(\bxi_i)) \notag\\
 &\qquad \qquad \leq 0
\end{align}
where we use the definition:
\begin{align}
\m_i^\dagger(\bxi_i) \triangleq \sum_{s^\prime \in \mathcal{S}}  \Big( \prod_{\ell =1}^K \! L_\ell(\bxi_{\ell,i} | s^\prime) \sum_{\ell \in \mathcal{N}_k}  c_{\ell k } \bmeta_{\ell,i} (s^\prime) \Big ) ,
\end{align}
which is a density (or mass function if observations are discrete) since:
\begin{align}
    \int_{\xi_i} \m_i^\dagger(\xi_i) d\xi_i &= \int_{\xi_i} \sum_{s^\prime \in \mathcal{S}}  \Big( \prod_{\ell =1}^K \! L_\ell(\bxi_{\ell,i} | s^\prime) \sum_{\ell \in \mathcal{N}_k}  c_{\ell k } \bmeta_{\ell,i} (s^\prime) \Big )  d\xi_i \notag \\
    &=  \sum_{s^\prime \in \mathcal{S}} \Big [ \underbrace{\int_{\xi_i} \prod_{\ell=1}^K L_\ell(\bxi_{\ell,i} | s^\prime)  d\xi_i}_{1} \sum_{\ell=1}^K c_{\ell k} \bmeta_{\ell,i} (s^\prime) \Big ]  \notag \\
    &= \sum_{s^\prime \in \mathcal{S}} \Big [  \sum_{\ell=1}^K c_{\ell k} \bmeta_{\ell,i} (s^\prime) \Big ] \notag \\
    &= \sum_{\ell=1}^K c_{\ell k}  \Big [ \sum_{s^\prime \in \mathcal{S}}    \bmeta_{\ell,i} (s^\prime) \Big ] = 1.
\end{align}
Notice that the expression in \eqref{eq:nus_before_defined} can be rewritten as
\begin{align}\label{eq:vi_difference_ours}
 &\e_{\f_i} \Bigg [\log \sum_{s^\prime \in \mathcal{S}} \!\! \Big( \prod_{\ell=1}^K\! (L_\ell(\bxi_{\ell,i} | s^\prime))^{\beta c_{\ell k}} \!\sum_{\ell \in \mathcal{N}_k} \! c_{\ell k } \bmeta_{\ell,i} (s^\prime) \Big ) \Bigg ] \notag\\
&\qquad - \e_{\f_i} \Bigg [\log \sum_{s^\prime \in \mathcal{S}} \!\! \Big( \prod_{\ell =1}^K\! L_\ell(\bxi_{\ell,i} | s^\prime) \!\sum_{\ell \in \mathcal{N}_k} \! c_{\ell k } \bmeta_{\ell,i} (s^\prime) \Big ) \Bigg ] \notag \\
&=\e_{\f_i} \Big [ f( \bvart_{k,i})  \Big ] -\e_{\f_i} \Big [ f (\widetilde{\bvart}_{k,i}) \Big ] ,
\end{align}
if we use the LogSumExp function $f$ from \eqref{eq:f_logsumexp_def} and use the definitions:
\begin{equation}\label{eq:vi_plus_def}
 \bvart_{k,i} \triangleq \mcl \Bigg \{ \log \Big( \prod_{\ell=1}^K\! (L_\ell(\bxi_{\ell,i} | s^\prime))^{\beta c_{\ell k}} \!\sum_{\ell \in \mathcal{N}_k} \! c_{\ell k } \bmeta_{\ell,i} (s^\prime) \Big ) \Bigg \}_{s^\prime \in \mathcal{S}}
\end{equation}
and
\begin{equation}\label{eq:vi_neg_def}
 \widetilde{\bvart}_{k,i} \triangleq \mcl \Bigg \{ \log  \Big( \prod_{\ell =1}^K\! L_\ell(\bxi_{\ell,i} | s^\prime) \!\sum_{\ell \in \mathcal{N}_k} \! c_{\ell k } \bmeta_{\ell,i} (s^\prime) \Big ) \Bigg \}_{s^\prime \in \mathcal{S}}.
\end{equation}
Following the steps in \eqref{eq:first_mean_val_1} and \eqref{eq:log_trans_difference}, this difference can be bounded as:
\begin{align}
&\e_{\f_i} \Big [ f( \bvart_{k,i})  \Big ] -\e_{\f_i} \Big [ f (\widetilde{\bvart}_{k,i}) \Big ] \notag \\
 &\leq\e_{\xi_i} \Bigg \| \mcl \Big \{  \sum_{\ell=1}^K (  \beta c_{\ell k}-1) \log L_\ell(\bxi_{\ell,i} | s^\prime)  \Big\}_{s^\prime \in \mathcal{S}} \Bigg \|_{\infty}.
\end{align}
Moreover, by assumptions on the graph topology  \eqref{eq:c_matrix_assumption} and on the likelihood functions \eqref{eq:log_bound_assumption}, this expression can be further bounded as \cite{kayaalp_dist_bayesian}:
\begin{align}
\Bigg \| \mcl \Big \{  \sum_{\ell=1}^K (  \beta c_{\ell k}-1) \log L_\ell(\bxi_{\ell,i} | s^\prime)  \Big\}_{s^\prime \in \mathcal{S}} \Bigg \|_{\infty} \leq \sqrt{K} \beta \lambda B \label{eq:normalization_term_bound_small}
\end{align}
Subsequently, if we insert the bounds \eqref{eq:first_4_term}, \eqref{eq:prior_wrt_risk}, and \eqref{eq:normalization_term_bound_small} to \eqref{eq:log_ratio_expansion}, we arrive at the bound on the risk function:
\begin{align}
J_{k,i}  &\leq  \e_{\xi_i,s_i} \Big[ \sum_{\ell=1}^K (1  - \beta c_{\ell k}) \log L_\ell(\bxi_{\ell,i} | \s_{i})  \Big] \notag \\
&\quad + \kappa (\bT) \sum_{\ell \in \mathcal{N}_k} c_{\ell k}  {J_{\ell,i-1}}  + \sqrt{K} \beta \lambda B + (K-d_{\textup{min}})\: \tau  \notag \\
&\stackrel{\eqref{eq:normalization_term_bound_small}}{\leq}   \kappa (\bT) \sum_{\ell \in \mathcal{N}_k} c_{\ell k}  {J_{\ell,i-1}} +2 \sqrt{K} \beta \lambda B + (K-d_{\textup{min}})\: \tau .\label{eq:combine_risk_mean}
\end{align}
Expanding this recursion over time yields:
\begin{align}
J_{k,i} &\leq (2\sqrt{K} \beta  \lambda B+ (K-d_{\textup{min}})\: \tau) \sum_{j=0}^{i-1} (\kappa (\bT))^{j}  \notag \\ &\qquad + (\kappa (\bT))^{i} \sum_{\ell =1}^K [C^{i}]_{\ell k} J_{\ell,0} \notag \\
&=    \frac{1-(\kappa (\bT))^{i}}{1-\kappa (\bT)}  (2\sqrt{K} \beta  \lambda B+ (K-d_{\textup{min}})\: \tau) \notag \\& \qquad + (\kappa (\bT))^{i} \sum_{\ell =1}^K [C^{i}]_{\ell k} J_{\ell,0} ,
\end{align}
which implies that if \( \kappa (\bT) < 1 \), the risk function is bounded as $i \to \infty$:
\begin{align}
\limsup_{i \rightarrow \infty} J_{k,i} \leq   \frac{2 \sqrt{K} \beta  \lambda B+(K-d_{\textup{min}})\: \tau}{1-\kappa (\bT)}  .
\end{align}
By \eqref{eq:prior_wrt_risk}, this also implies that
\begin{align}
 \limsup_{i \rightarrow \infty} \widetilde{J}_{k,i} &\leq  (K-d_{\textup{min}})\tau + \kappa (\bT)  \limsup_{i \rightarrow \infty} J_{k,i} \notag \\& \leq \frac{(K-d_{\textup{min}})\tau}{1-\kappa (\bT)} + \kappa (\bT) \frac{2 \sqrt{K} \beta  \lambda B}{1-\kappa (\bT)}  .
\end{align}

\subsection{Proof of Corollary~\ref{cor:pinsker}}\label{appendix:risk_corollary}

In view of the Bretagnolle-Huber inequality \cite{bretagnolle_huber}, it holds that
\begin{equation}
    \sum_{s \in \mathcal{S}} \big |\bmu_i(s) - \bmu_{k,i}(s) \big | \leq 2\big (  1-\exp \{ - \dkl (\bmu_i || \bmu_{k,i}) \} \big )^{\frac{1}{2}}  .
    \end{equation}
If we take the expectation of both sides, we get:
\begin{align}
     \e \Big [ \!\!\sum_{s \in \mathcal{S}}\! \big |\bmu_i(s) - \bmu_{k,i}(s) \big | \Big ] \!\!&\leq  2\e \big (   1-\exp \{ - \dkl (\bmu_i || \bmu_{k,i}) \} \big )^{\frac{1}{2}} \notag \\
     &\stackrel{(a)}{\leq}  2 \big (   1-\e \exp \{ - \dkl (\bmu_i || \bmu_{k,i}) \} \big )^{\frac{1}{2}} \notag \\
     &\stackrel{(b)}{\leq}  2 \big (   1-\exp \{ - J_{k,i} \} \big )^{\frac{1}{2}},
\end{align}
where \( (a) \) and \( (b) \) follow from Jensen's inequality. Together with Theorem~\ref{th:kl_without_net_assumption}, this implies that
\begin{equation}
      \e \big \|\bmu_i - \bmu_{k,i} \big \|_1 \leq  \btv ,    
\end{equation}
where we use the definition \eqref{eq:btv_definition}. Furthermore, on account of the fact that \( \ell_2\) norm is no greater than \(\ell_1\) norm in \( \mathbb{R}^S\), it is also true that
\begin{equation}
       \e \big \|\bmu_i - \bmu_{k,i} \big \| \leq \btv.
\end{equation}
With similar arguments, it can be shown that
\begin{equation}
    \e \big \|\bmeta_i - \bmeta_{k,i} \big \| \leq \wbtv,
\end{equation}
where we use the definition \eqref{eq:wbtv_definition}.
\subsection{Proof of Lemma~\ref{lemma:h_difference}}\label{sec:appendix_h_difference}
Inserting the definitions \eqref{eq:hki_definition} and \eqref{eq:histar_definition}, the expected difference can be expanded as
\begin{align}\label{eq:proof_delta_two_terms}
     \e \|\bH_{k,i}- \bH_i^\star\|\!&= \e \Big \|\phi(\bmu_{k,i})\phi(\bmu_{k,i})^{\T} -\gamma \phi(\bmu_{k,i})\phi(\bmeta_{k,i+1})^{\T}\notag \\ &\qquad -   \phi(\bmu_{i})\phi(\bmu_{i})^{\T}+\gamma \phi(\bmu_{i})\phi(\bmeta_{i+1})^{\T} \Big \| \notag \\
      &\leq \e \Big  \|\phi(\bmu_{k,i})\phi(\bmu_{k,i})^{\T} - \phi(\bmu_{i})\phi(\bmu_{i})^{\T} \Big \| \notag \\
      & +\! \gamma \e \Big \|\phi(\bmu_{k,i})\phi(\bmeta_{k,i+1})^{\T} -  \phi(\bmu_{i})\phi(\bmeta_{i+1})^{\T} \Big \|,
\end{align}
where the last step follows from the triangle inequality. Here, the first term can be bounded as
\begin{align}\label{eq:delta_first_bound}
    \Big \|\phi(&\bmu_{k,i})\phi(\bmu_{k,i})^{\T} - \phi(\bmu_{i})\phi(\bmu_{i})^{\T} \Big \| \notag \\
    &\quad \leq \Big \|\phi(\bmu_{k,i})(\phi(\bmu_{k,i})^{\T} - \phi(\bmu_{i})^{\T}) \Big \| \notag \\
    &\qquad + \Big \|(\phi(\bmu_{k,i}) - \phi(\bmu_{i}))\phi(\bmu_{i})^{\T} \Big \| \notag \\
     &\quad \leq \big \|\phi(\bmu_{k,i}) \big \| \big \| \phi(\bmu_{k,i}) - \phi(\bmu_{i}) \big \| \notag \\
    &\qquad + \big \|\phi(\bmu_{k,i}) - \phi(\bmu_{i}) \big \| \big \| \phi(\bmu_{i}) \big \| \notag \\
    & \quad \stackrel{(a)}{\leq} \bfi \lfi \|\bmu_{k,i} - \bmu_{i} \| + \bfi \lfi \|\bmu_{k,i} - \bmu_{i} \| ,
    \end{align}
   where \( (a) \) follows from  Assumption~\ref{assumption:feature}. Taking expectations and using \eqref{eq:btv_eq} and \eqref{eq:delta_first_bound}, it follows that
\begin{equation}\label{eq:delta_first_bound_ex}
      \e \Big \|\phi(\bmu_{k,i})\phi(\bmu_{k,i})^{\T} - \phi(\bmu_{i})\phi(\bmu_{i})^{\T} \Big \| \leq2 \bfi \lfi \btv.
\end{equation}    
Similarly, the second term in \eqref{eq:proof_delta_two_terms} can be bounded as
\begin{align}\label{eq:delta_second_bound}
     \Big \|\phi&(\bmu_{k,i})\phi(\bmeta_{k,i+1})^{\T} -  \phi(\bmu_{i})\phi(\bmeta_{i+1})^{\T} \Big \| \notag \\
     &\leq \Big \|\phi(\bmu_{k,i})(\phi(\bmeta_{k,i+1})^{\T} - \phi(\bmeta_{i+1})^{\T}) \Big \| \notag \\
     &\quad + \Big \|(\phi(\bmu_{k,i})-  \phi(\bmu_{i}))\phi(\bmeta_{i+1})^{\T} \Big \| \notag \\
    &\leq \big \|\phi(\bmu_{k,i}) \big \| \big \| \phi(\bmeta_{k,i+1}) - \phi(\bmeta_{i+1}) \big \| \notag \\
     &\quad + \big \|\phi(\bmu_{k,i})-  \phi(\bmu_{i}) \big \| \big \| \phi(\bmeta_{i+1}) \big \| \notag \\
      & \stackrel{(a)}{\leq} \bfi \lfi \|\bmeta_{k,i+1} - \bmeta_{i+1} \| + \bfi \lfi \|\bmu_{k,i} - \bmu_{i} \| 
  \end{align} 
  where \( (a) \) follows from Assumption~\ref{assumption:feature}. Using \eqref{eq:btv_eq} and \eqref{eq:btv_kappa_eq} we get:
      \begin{align}\label{eq:delta_second_bound_ex}
        \e \Big \|\phi(\bmu_{k,i})\phi(\bmeta_{k,i+1})^{\T} &-  \phi(\bmu_{i})\phi(\bmeta_{i+1})^{\T} \Big \|  \notag \\&\leq \bfi \lfi  (\btv + \wbtv ).
      \end{align}
Combining \eqref{eq:delta_first_bound_ex} and \eqref{eq:delta_second_bound_ex} in addition to the fact that $\wbtv~\leq~\btv$ (since \(\kappa (\bT) < 1\)) yields:
\begin{equation}
  \e  \|\bH_{k,i}- \bH_i^\star\| \leq 2\bfi \lfi \btv(1+ \gamma ).
\end{equation}

\subsection{Proof of Theorem \ref{th:network_disagreement}}\label{sec:appendix_network_disagreement}

For compactness of notation, it is useful to introduce the following quantities, which collect variables from across all agents:
\begin{align}
  \bcw_{i} &\triangleq \mathrm{col} \left \{ \w_{1,i}, \ldots, \w_{K,i} \right \} \\
  \mathcal{C} &\triangleq C \otimes I_M\\
 \bmch_i &\triangleq \mathrm{diag} \left \{ \bH_{k,i}\right \}_{k=1}^K \\
 \bmch_i^\star &\triangleq I_K \otimes \bH_i^\star \\
  \bD_i &\triangleq \mathrm{col} \left \{ \bD_{k,i}\right \}_{k=1}^K
\end{align}
Then, the equations \eqref{eq:delta_parameter}--\eqref{eq:combine_parameter} can be written as:
\begin{equation}\label{eq:wi_cal_recursion}
	\bcw_{i+1} = \mathcal{C}^{\T} \Big( \left(I(1-2\alpha\rho)-\alpha \bmch_i \right) \bcw_{i} + \alpha \bD_i \Big).
\end{equation}
Moreover, we can define the following \(K\)-times extended centroid vector:
\begin{equation}\label{eq:def_centroid_vec}
	\bcw_{c, i} \triangleq \mathds{1}_K \otimes \w_{c, i} = \left( \frac{1}{K} \mathds{1}_K \mathds{1}_K^{\T} \otimes I \right) \bcw_i.
\end{equation}
If we decompose \( \bmch_i \) into its centralized component \( \bmch_i^\star \) and the respective disagreement matrix \( \bm{\Delta}_i \triangleq \bmch_i - \bmch_i^\star\), we obtain:
\begin{align}\label{eq:network_disagreement_recursion}
    	&\bcw_{i+1} - \bcw_{c, i+1} \notag \\
	&=\left( \mathcal{C}^{\T} - \frac{1}{K} \mathds{1}_K \mathds{1}_K^{\T} \otimes I \right)\Big( \left(I(1-2\alpha\rho)-\alpha \bmch_i \right) \bcw_{i} + \alpha \bD_i \Big) \notag \\
	&=\left( \mathcal{C}^{\T} - \frac{1}{K} \mathds{1}_K \mathds{1}_K^{\T} \otimes I \right) \notag \\ &\qquad \Big( \left(I(1-2\alpha\rho)-\alpha \bmch_i^\star -\alpha \bm{\Delta}_i \right) \bcw_{i} + \alpha \bD_i \Big) \notag \\
	&= \left( \mathcal{C}^{\T} - \frac{1}{K} \mathds{1}_K \mathds{1}_K^{\T} \otimes I \right) \notag \\ &\qquad \Big( \left(I(1-2\alpha\rho)-\alpha \bmch_i^\star \right) (\bcw_{i}-\bcw_{c,i}) -\alpha \bm{\Delta}_i  \bcw_{i} + \alpha \bD_i \Big),
\end{align}
where the last step follows from the fact that
\begin{align}
  &\mathcal{C}^{\T}  \big(I(1-2\alpha\rho)-\alpha \bmch_i^\star \big) \bcw_{c,i} \notag \\ & \quad = \Big (\frac{1}{K} \mathds{1}_K \mathds{1}_K^{\T} \otimes I  \Big ) \big(I(1-2\alpha\rho)-\alpha \bmch_i^\star \big) \bcw_{c,i}.
\end{align}
Furthermore, taking the norms of both sides in \eqref{eq:network_disagreement_recursion} leads to
\begin{align}\label{eq:wi_wci_norm}
    	& \Big \| \bcw_{i+1} - \bcw_{c, i+1} \Big \| \notag \\
    	& \leq \Big \| \mathcal{C}^{\T} - \frac{1}{K} \mathds{1}_K \mathds{1}_K^{\T} \otimes I  \Big \| \notag \\ &\qquad \Big \| \left(I(1-2\alpha\rho)-\alpha \bmch_i^\star \right) (\bcw_{i}-\bcw_{c,i}) -\alpha \bm{\Delta}_i  \bcw_{i} + \alpha \bD_i \Big \| \notag \\
    	& \leq \! \Big \| \mathcal{C}^{\T} - \frac{1}{K} \mathds{1}_K \mathds{1}_K^{\T} \otimes I  \Big \| \left  \| I(1-2\alpha\rho)-\alpha \bmch_i^\star \right \| \! \| \bcw_{i}-\bcw_{c,i} \| \notag \\ & +  \alpha \Big \| \mathcal{C}^{\T} - \frac{1}{K} \mathds{1}_K \mathds{1}_K^{\T} \otimes I  \Big \| \Big (  \| \bm{\Delta}_i \| \| \bcw_{i} \| + \| \bD_i \| \Big ) .
\end{align}
Since the combination matrix \( C \) is a primitive stochastic matrix, it follows from the Perron-Frobenius theorem \cite{pillai2005,sayed_2014} that its maximum eigenvalue is 1, and all other eigenvalues are strictly smaller than 1 in absolute value. Moreover, $C$ is assumed to be symmetric, therefore its eigenvalue decomposition has the following form:
\begin{align*}
C &= U \Lambda U^\top \notag \\ &= \begin{bmatrix} u_1 & u_2 & \cdots & u_K \end{bmatrix} \begin{bmatrix} 1 & 0 & \cdots & 0 \\ 0 & \lambda_2 & \cdots & 0 \\ \vdots & \vdots & \ddots & \vdots \\ 0 & 0 & \cdots & \lambda_K \end{bmatrix} \begin{bmatrix} u_1^\top \\ u_2^\top \\ \vdots \\ u_K^\top \end{bmatrix}
\end{align*}
where $U$ is the orthogonal matrix of eigenvectors $\{u_k\}$, and $\Lambda$ is the diagonal matrix of eigenvalues. Additionally, the powers of $C$ converge (because it is primitive) to the scaled all-ones matrix (because it is doubly-stochastic):
\begin{align*}
\lim_{i \to \infty} C^i &= \begin{bmatrix} u_1 & u_2 & \cdots & u_K \end{bmatrix} \begin{bmatrix} 1 & 0 & \cdots & 0 \\ 0 & 0 & \cdots & 0 \\ \vdots & \vdots & \ddots & \vdots \\ 0 & 0 & \cdots & 0 \end{bmatrix} \begin{bmatrix} u_1^\top \\ u_2^\top \\ \vdots \\ u_K^\top \end{bmatrix} \notag \\&= \frac{1}{K}\begin{bmatrix} 1 & 1 & \cdots & 1 \\ 1 & 1 & \cdots & 1 \\ \vdots & \vdots & \ddots & \vdots \\ 1 & 1 & \cdots & 1 \end{bmatrix} = \frac{1}{K} \mathds{1}_K \mathds{1}_K^\top
\end{align*}
Therefore, the difference of these matrices becomes:
\begin{align}
    &C-\frac{1}{K} \mathds{1}_K \mathds{1}_K^\top = U \begin{bmatrix} 0 & 0 & \cdots & 0 \\ 0 & \lambda_2 & \cdots & 0 \\ \vdots & \vdots & \ddots & \vdots \\ 0 & 0 & \cdots & \lambda_K \end{bmatrix}  U^\top,
\end{align}
which implies:
\begin{align}    
    \Big \| C-\frac{1}{K} \mathds{1}_K \mathds{1}_K^\top \Big \| = \lambda_2
\end{align}
where $\lambda_2$ is the second largest modulus eigenvalue of $C$. Moreover, the Kronecker product with the identity matrix does not change the spectral norm, hence:
\begin{equation}
    \Big \| \mathcal{C}^{\T} - \frac{1}{K} \mathds{1}_K \mathds{1}_K^{\T} \otimes I  \Big \| =\lambda_2 < 1.
\end{equation}
Moreover, we know from Lemma~\ref{lemma:h_difference} that
\begin{equation}
   \e \| \bm{\Delta}_i \| \leq  2\bfi \lfi \btv(1+\gamma).
\end{equation}
Additionally, in Appendix~\ref{sec:auxiliary_results_section}, we establish \eqref{eq:nd_proof_additional_1}--\eqref{eq:nd_proof_additional_4} which hold for any realization (with probability one). From \eqref{eq:rho_norm_bound}, note that:
\begin{equation}\label{eq:nd_proof_additional_1}
    \bigg \| I(1-2\alpha\rho)-\alpha \bmch_i^\star \bigg \| < 1
\end{equation}
whenever \( \rho > \gamma \lfi \bfi/ \sqrt{2} \). Specifically, if \( \rho \geq 0.75 \gamma \lfi \bfi  \), then
\begin{equation}\label{eq:nd_proof_additional_2}
    \bigg \| I(1-2\alpha\rho)-\alpha \bmch_i^\star \bigg \| \leq (1-0.08 \alpha \gamma \lfi \bfi ).
\end{equation}
In addition, we show in Lemma~\ref{lemma:bounded_w} that
\begin{align}\label{eq:nd_proof_additional_3}
    \left\| \bcw_{i} \right\| \leq  \frac{\sqrt{K}  \rma}{0.08 \gamma \lfi }  
\end{align}
and in expression \eqref{eq:dki_bound} that
\begin{equation}\label{eq:nd_proof_additional_4}
    \| \bD_i \|  \leq \sqrt{K} \rma \bfi.
\end{equation}
Inserting these results into \eqref{eq:wi_wci_norm} yields the following norm recursion:
\begin{align}\label{eq:simple_wi_wci_norm}
    &\e \big \| \bcw_{i+1} - \bcw_{c, i+1} \big \| \notag \\& \leq \lambda_2 (1-0.08\alpha \gamma \bfi \lfi ) \e \big \| \bcw_{i} - \bcw_{c, i} \big \| + \alpha \lambda_2 \sqrt{K} \epsilon.
\end{align}
Let us define the constant \( \widetilde{\lambda}_2 \triangleq \lambda_2  (1-0.08\alpha \gamma \bfi \lfi )  \). Iterating \eqref{eq:simple_wi_wci_norm} over time, we arrive at
\begin{align}
	& \e \left \| \bcw_{i+1} - \bcw_{c, i+1} \right \| \notag \\
	&\qquad\leq\: \widetilde{\lambda}_2^{i+1} \|\cw_0 - \cw_{c, 0}\| +   \alpha \lambda_2 \sqrt{K} \epsilon \sum_{j = 1}^{i+1} \widetilde{\lambda}_2^{i+1-j} \notag \\
	&\qquad\leq\: \widetilde{\lambda}_2^{i+1} \|\cw_0 - \cw_{c, 0}\| + \alpha \lambda_2 \sqrt{K} \epsilon \frac{1}{1-\widetilde{\lambda}_2} \notag \\
	&\qquad\stackrel{(a)}{\leq}\: \alpha \lambda_2 \sqrt{K} \epsilon \frac{1}{1-\widetilde{\lambda}_2} + O(\alpha^2)
\end{align}
where \( (a) \) holds whenever:
\begin{align}
    &\widetilde{\lambda}_2^i \|\cw_0 - \cw_{c, 0}\| \leq c \alpha^2 \notag \\& \Longleftrightarrow i \log \widetilde{\lambda}_2 \leq 2 \log \alpha + \log c - \log \|\cw_0 - \cw_{c, 0}\| \notag \\
  &\Longleftrightarrow i \geq \frac{2\log \alpha }{\log \widetilde{\lambda}_2}+O(1)=O(\log \alpha)=o(1/\alpha),
\end{align}
where \( c \) is an arbitrary constant.
\subsection{Proof of Theorem~\ref{th:centralized_disagreement}}\label{sec:appendix_centralized_dif}
We begin by rewriting the baseline strategy recursion \eqref{eq:cent_pol_learn_del_baseline}--\eqref{eq:cent_pol_learn_w_baseline} in the form:
\begin{align}\label{eq:centralized_recursion_no_ext}
     \w_{i+1}^\star = \left((1-2\rho\alpha)I-\alpha \bH_{i}^\star\right)\w_{i}^\star + \alpha \bD_{i}^\star,
\end{align}
where \( \bH_{i}^\star \) is defined in \eqref{eq:histar_definition}, and
\begin{equation}
    \bD_{i}^\star \triangleq \Big (\frac{1}{K} \sum_{k=1}^K \rb_{k,i} \Big) \phi(\bmu_{i}).
\end{equation}
We introduce the \( K\)-times extended versions of the vectors:
\begin{equation}
     \bm{\mathcal{D}}_i^\star \triangleq \mathds{1}_K \otimes \bD_i^\star, \quad  \bcw_i^\star =  \mathds{1}_K \otimes  \w_{i}^\star.
\end{equation}
Then, the baseline recursion \eqref{eq:centralized_recursion_no_ext} transforms into
\begin{align}\label{eq:cal_w_star_recursion}
     \bcw_{i+1}^\star = \left((1-2\rho\alpha)I-\alpha \bmch_{i}^\star\right)\bcw_{i}^\star + \alpha \bm{\mathcal{D}}_i^\star.
\end{align}
It follows from the extended network centroid definition \eqref{eq:def_centroid_vec} and \eqref{eq:cal_w_star_recursion} that
\begin{align}\label{eq:w_i_star_w_c_recursion}
    &\bcw_{i+1}^\star -  \bcw_{c,i+1} \notag \\ &= \left(I(1-2\alpha\rho)-\alpha \bmch_i^\star \right) (\bcw_{i}^\star-\bcw_{c,i}) \notag \\& -\alpha \bigg (\frac{1}{K} \mathds{1}_K \mathds{1}_K^{\T} \otimes I \bigg ) \bm{\Delta}_i  \bcw_{i} + \alpha \bigg (\frac{1}{K} \mathds{1}_K \mathds{1}_K^{\T} \otimes I \bigg ) (\bm{\mathcal{D}}_i^\star-\bD_i)
\end{align}
where we used the facts that
\begin{align}
     \Big (\frac{1}{K} \mathds{1}_K \mathds{1}_K^{\T} \otimes I \Big )  \bm{\mathcal{D}}_i^\star=  \bm{\mathcal{D}}_i^\star,
\end{align}
and
\begin{equation}
     \Big (\frac{1}{K} \mathds{1}_K \mathds{1}_K^{\T} \otimes I \Big ) \bmch_{i}^\star \bcw_{i} = \bmch_{i}^\star \bcw_{c,i}.
\end{equation}
Next, if we define the following average agent disagreement relative to the baseline term
\begin{equation}
    \widetilde{\bD}_i \triangleq \frac{1}{K} \sum_{k=1}^K (\bD_i^\star-\bD_{k,i}),
\end{equation}
it holds that
\begin{equation}
    \widetilde{\bm{\mathcal{D}}}_i \triangleq \mathds{1}_K \otimes \widetilde{\bD}_i = \Big (\frac{1}{K} \mathds{1}_K \mathds{1}_K^{\T} \otimes I \Big ) (\bm{\mathcal{D}}_i^\star-\bD_i).
\end{equation}
Subsequently, taking the norm of both sides in \eqref{eq:w_i_star_w_c_recursion} and applying the triangle inequality, we get
\begin{align}\label{eq:w_i_star_w_c_norm}
    & \Big \| \bcw_{i+1}^\star -  \bcw_{c,i+1} \Big \| \notag \\ &\quad\leq \Big \|I(1-2\alpha\rho)-\alpha \bmch_i^\star \Big \| \Big \| \bcw_{i}^\star-\bcw_{c,i} \Big \| \notag \\&\qquad +\alpha \Big \| \frac{1}{K} \mathds{1}_K \mathds{1}_K^{\T} \otimes I \Big \| \big \| \bm{\Delta}_i \big \| \big \| \bcw_{i} \big \| + \alpha \big \| \widetilde{\bm{\mathcal{D}}}_i \big \|.
\end{align}
First, observe that 
\begin{align}
    \Bigg \| \frac{1}{K} \mathds{1}_K \mathds{1}_K^{\T} \otimes I \Bigg \| = 1.
\end{align}
Moreover, from Assumption~\ref{assumption:feature} and Corollary~\ref{cor:pinsker}, it holds that
\begin{align}
   \e \big \| \widetilde{\bD}_i \big \| &= \e \Big \| \frac{1}{K} \sum_{k=1}^K \rb_{k,i} (\phi(\bmu_{i})-\phi(\bmu_{k,i})) \Big \| \notag \\
    &\leq \rma \lfi \btv,
\end{align}
and accordingly,
\begin{equation}
    \e \big \| \widetilde{\bm{\mathcal{D}}}_i \big \| \leq \sqrt{K} \rma \lfi \btv.
\end{equation}
By using the same bounds \eqref{eq:nd_proof_additional_1}--\eqref{eq:nd_proof_additional_4} from Appendix~\ref{sec:appendix_network_disagreement} for the other terms (which are established in Lemma~\ref{lemma:h_difference}, Lemma~\ref{lemma:bounded_w}, \eqref{eq:rho_norm_bound}, and \eqref{eq:dki_bound}), we arrive at the recursion:
\begin{align}
     & \e \Big \| \bcw_{i+1}^\star -  \bcw_{c,i+1} \Big \| \notag \\& \qquad \leq (1-0.08\alpha \gamma \bfi \lfi) \e \big \| \bcw_{i}^\star - \bcw_{c, i} \big \| + \alpha \sqrt{K} \epsilon^\star,
\end{align}
where
    \begin{equation}
    \epsilon^\star \triangleq \rma \btv \Big (  \frac{2\bfi (1+ \gamma  )}{0.08 \gamma}  +   \lfi \Big )  .
\end{equation}
Iterating over time, we get:
\begin{align}
	& \e \Big \| \bcw_{i+1}^\star -  \bcw_{c,i+1} \Big \| \notag \\
	&\quad\leq\: (1-0.08\alpha \gamma \bfi \lfi)^{i+1} \|\cw_0^\star - \cw_{c, 0}\| \notag \\&\qquad+   \alpha  \sqrt{K} \epsilon^\star \sum_{j = 1}^{i+1} (1-0.08\alpha \gamma \bfi \lfi)^{i+1-j} \notag \\
	&\quad\leq\: (1-0.08\alpha \gamma \bfi \lfi)^{i+1} \|\cw_0^\star - \cw_{c, 0}\| +    \frac{\sqrt{K} \epsilon^\star}{0.08 \gamma \bfi \lfi } \notag \\
	&\quad\stackrel{(a)}{\leq}\: \frac{\sqrt{K} \epsilon^\star}{0.08 \gamma \bfi \lfi } + o(1)
\end{align}
where \( (a) \) holds whenever
\begin{align}
    &(1-0.08\alpha \gamma \bfi \lfi)^{i+1}  \big\|\cw_{0}^\star - \cw_{c,0}\big\| = o(1)  \notag \\&\qquad \Longleftrightarrow i \log (1-0.08\alpha \gamma \bfi \lfi) = o(1) \notag \\
  &\qquad\Longleftrightarrow i \geq \frac{o(1)}{\log (1-0.08\alpha \gamma \bfi \lfi)} \geq o\left(\frac{1}{\alpha \gamma \bfi \lfi}\right).
\end{align}

\subsection{Auxiliary Results}\label{sec:auxiliary_results_section}
In the following lemma, we prove that the value function parameters are bounded in norm.
\begin{lemma}[{\bf Bounded parameters}]\label{lemma:bounded_w}For each agent \( k \in \mathcal{K} \), the iterate \( \w_{k,i} \) is bounded in norm if \( \rho > \gamma \bfi \lfi / \sqrt{2}  \), with probability one. In particular, if \( \rho \geq  0.75 \gamma \bfi \lfi \), then
\begin{align}
    \left\| \bcw_{i} \right\| \leq  \frac{\sqrt{K}  \rma}{0.08 \gamma \lfi }  
\end{align}
after \(i\geq i_0=  o\left(1/(\alpha \gamma\bfi \lfi )\right) \) iterations. \\

\begin{myproof}
\normalfont
Taking the norms of both sides of \eqref{eq:wi_cal_recursion} yields:
\begin{align}\label{eq:norm_recursion}
   \big \|	\bcw_{i+1} \big \| &= \Big \| \mathcal{C}^{\T} \big( \left((1-2\alpha\rho)I-\alpha \bmch_i \right) \bcw_{i} + \alpha \bD_i \big) \Big \| \notag \\
   &\leq \Big \| \mathcal{C}^{\T} \Big \| \Big \|  \left((1-2\alpha\rho)I-\alpha \bmch_i \right) \bcw_{i} + \alpha \bD_i  \Big \| \notag \\
   &\stackrel{(a)}{\leq} \Big \|   \left((1-2\alpha\rho)I-\alpha \bmch_i \right) \bcw_{i} + \alpha \bD_i  \Big \| \notag \\
      &\leq \big \|   (1-2\alpha\rho)I-\alpha \bmch_i  \big \| \big \| \bcw_{i} \big \| + \alpha \big \| \bD_i  \big \| 
\end{align}
where \( (a) \) follows from the fact that the singular values of doubly-stochastic matrices are equal to one. Note that
\begin{align}\label{eq:rho_norm_bound}
     &\big \|   (1-2\alpha\rho)I-\alpha \bmch_{k,i}  \big \| \notag \\ & =   \big \|   (1-2\alpha\rho)I-\alpha  \phi(\bmu_{k,i})\phi(\bmu_{k,i})^{\T} +\alpha \gamma  \phi(\bmu_{k,i})\phi(\bmeta_{k,i+1})^{\T}   \big \| \notag \\
     &=   \big \|   (1-2\alpha\rho)I-\alpha (1-\gamma)  \phi(\bmu_{k,i})\phi(\bmu_{k,i})^{\T} \notag \\& \qquad \qquad -\alpha \gamma  \phi(\bmu_{k,i})\big (\phi(\bmu_{k,i})^{\T}-\phi(\bmeta_{k,i+1})^{\T}  \big ) \big \| \notag \\
          &\leq \big \|   (1-2\alpha\rho)I-\alpha (1-\gamma)  \phi(\bmu_{k,i})\phi(\bmu_{k,i})^{\T} \big \| \notag \\
     &\qquad \qquad +\alpha \gamma \| \phi(\bmu_{k,i}) \| \big \| \phi(\bmu_{k,i})^{\T}-\phi(\bmeta_{k,i+1})^{\T} \big \| \notag \\
     &\stackrel{(a)}{\leq} (1-2\alpha\rho) +\alpha \gamma \| \phi(\bmu_{k,i}) \| \big \| \phi(\bmu_{k,i})^{\T}-\phi(\bmeta_{k,i+1})^{\T} \big \| \notag \\
     &\stackrel{(b)}{\leq} (1-2\alpha\rho) + \alpha \gamma \bfi \lfi \big \| \bmu_{k,i}-\bmeta_{k,i+1} \big \| \notag \\
     &\stackrel{(c)}{\leq} (1-2\alpha\rho) + \alpha \gamma \bfi \lfi \sqrt{2}
\end{align}
where \( (a)\) follows from the equality of spectral norm and maximum eigenvalue for symmetric matrices, \( (b)\) follows from Assumption~\ref{assumption:feature}, and \( (c) \) follows from the fact that the mean-square distance cannot exceed 2 over the probability simplex. The upper bound in \eqref{eq:rho_norm_bound} is smaller than 1 whenever \( \rho > \gamma \bfi \lfi / \sqrt{2}  \). Moreover,
\begin{align}\label{eq:dki_bound}
    \big \| \bD_{k,i}  \big \| =   \big \|  \rb_{k,i} \phi(\bmu_{k,i}) \big \| \leq \rma \bfi.
\end{align}
As a result, if \( \rho \geq 0.75 \gamma \bfi \lfi   \), we get:
\begin{align}
      \big \|	\bcw_{i+1} \big \| &\stackrel{\eqref{eq:rho_norm_bound}}{\leq}  (1- 0.08 \alpha \gamma \bfi \lfi)  \big\|\bcw_{i}\big\| + \alpha \big \| \bD_i  \big \| \notag \\
      &\stackrel{\eqref{eq:dki_bound}}{\leq}  (1-0.08 \alpha \gamma \bfi \lfi )  \big\|\bcw_{i}\big\| + \alpha \sqrt{K} \rma \bfi.
\end{align}
Iterating this recursion starting from \( i = 0 \) results in
\begin{align}
     \big \|\bcw_{i+1} \big \| &\leq \alpha \sqrt{K} \rma \bfi \sum_{j = 1}^{i+1} (1-0.08 \alpha \gamma \bfi \lfi)^{i+1-j} \notag \\ & \qquad + (1-0.08 \alpha \gamma \bfi \lfi)^{i+1}  \big\|\cw_{0}\big\| \notag \\
     &\leq \frac{\sqrt{K} \rma}{0.08  \gamma  \lfi}   + (1-0.08 \alpha \gamma \bfi \lfi)^{i+1}  \big\|\bcw_{0}\big\| \notag \\
     &= \frac{\sqrt{K} \rma}{0.08  \gamma  \lfi} + o(1),
\end{align}
where the last step holds whenever
\begin{align}
    &(1-0.08 \alpha \gamma \bfi \lfi)^{i+1}  \big\|\cw_{0}\big\| = o(1)  \notag \\&\qquad \Longleftrightarrow i \log (1-0.08 \alpha \gamma \bfi \lfi) = o(1) \notag \\
  &\qquad\Longleftrightarrow i \geq \frac{o(1)}{\log (1-0.08 \alpha \gamma \bfi \lfi)} \geq o\left(\frac{1}{\alpha \gamma \bfi \lfi}\right).
\end{align}
\end{myproof}
\end{lemma}

\section*{ACKNOWLEDGMENTS}

The authors would like to thank Visa Koivunen for useful discussions on Dec-POMDPs and their applications in sensing.

\bibliographystyle{IEEEtran}
\bibliography{IEEEfull,IEEEexample}

\begin{thebibliography}{10}
\providecommand{\url}[1]{#1}
\csname url@rmstyle\endcsname
\providecommand{\newblock}{\relax}
\providecommand{\bibinfo}[2]{#2}
\providecommand\BIBentrySTDinterwordspacing{\spaceskip=0pt\relax}
\providecommand\BIBentryALTinterwordstretchfactor{4}
\providecommand\BIBentryALTinterwordspacing{\spaceskip=\fontdimen2\font plus
\BIBentryALTinterwordstretchfactor\fontdimen3\font minus
  \fontdimen4\font\relax}
\providecommand\BIBforeignlanguage[2]{{%
\expandafter\ifx\csname l@#1\endcsname\relax
\typeout{** WARNING: IEEEtran.bst: No hyphenation pattern has been}%
\typeout{** loaded for the language `#1'. Using the pattern for}%
\typeout{** the default language instead.}%
\else
\language=\csname l@#1\endcsname
\fi
#2}}

\bibitem{busoniu2008comprehensive}
L.~Busoniu, R.~Babuska, and B.~De~Schutter, ``A comprehensive survey of
  multiagent reinforcement learning,'' \emph{IEEE Transactions on Systems, Man,
  and Cybernetics, Part C (Applications and Reviews)}, vol.~38, no.~2, pp.
  156--172, 2008.

\bibitem{zhang2021multi}
K.~Zhang, Z.~Yang, and T.~Ba{\c{s}}ar, ``Multi-agent reinforcement learning: A
  selective overview of theories and algorithms,'' \emph{Handbook of
  Reinforcement Learning and Control}, pp. 321--384, 2021.

\bibitem{huang2010distributed}
J.~W. Huang, Q.~Zhu, V.~Krishnamurthy, and T.~Basar, ``Distributed correlated
  {Q}-learning for dynamic transmission control of sensor networks,'' in
  \emph{Proc. IEEE International Conference on Acoustics, Speech and Signal
  Processing}, 2010, pp. 1982--1985.

\bibitem{lunden2013multiagent}
J.~Lunden, S.~R. Kulkarni, V.~Koivunen, and H.~V. Poor, ``Multiagent
  reinforcement learning based spectrum sensing policies for cognitive radio
  networks,'' \emph{IEEE Journal of Selected Topics in Signal Processing},
  vol.~7, no.~5, pp. 858--868, 2013.

\bibitem{bhattacharya2021multiagent}
S.~Bhattacharya, S.~Kailas, S.~Badyal, S.~Gil, and D.~Bertsekas, ``Multiagent
  rollout and policy iteration for {POMDP} with application to multi-robot
  repair problems,'' in \emph{Proc. Conference on Robot Learning}.\hskip 1em
  plus 0.5em minus 0.4em\relax PMLR, 2021, pp. 1814--1828.

\bibitem{vinyals2019grandmaster}
O.~Vinyals, I.~Babuschkin, W.~M. Czarnecki, M.~Mathieu, A.~Dudzik, J.~Chung,
  D.~H. Choi, R.~Powell, T.~Ewalds, P.~Georgiev, \emph{et~al.}, ``Grandmaster
  level in {StarCraft II} using multi-agent reinforcement learning,''
  \emph{Nature}, vol. 575, no. 7782, pp. 350--354, 2019.

\bibitem{samvelyan2019}
M.~Samvelyan, T.~Rashid, C.~Schroeder~de Witt, G.~Farquhar, N.~Nardelli,
  T.~G.~J. Rudner, C.-M. Hung, P.~H.~S. Torr, J.~Foerster, and S.~Whiteson,
  ``The starcraft multi-agent challenge,'' in \emph{Proc. International
  Conference on Autonomous Agents and MultiAgent Systems}, ser. AAMAS '19,
  2019, p. 2186–2188.

\bibitem{lecun2015deep}
Y.~LeCun, Y.~Bengio, and G.~Hinton, ``Deep learning,'' \emph{Nature}, vol. 521,
  no. 7553, pp. 436--444, 2015.

\bibitem{oliehoek2016concise}
F.~A. Oliehoek and C.~Amato, \emph{A Concise Introduction to Decentralized
  {POMDPs}}.\hskip 1em plus 0.5em minus 0.4em\relax Springer, 2016.

\bibitem{sondik1978optimal}
E.~J. Sondik, ``The optimal control of partially observable {Markov} processes
  over the infinite horizon: Discounted costs,'' \emph{Operations Research},
  vol.~26, no.~2, pp. 282--304, 1978.

\bibitem{kaelbling1998}
L.~P. Kaelbling, M.~L. Littman, and A.~R. Cassandra, ``Planning and acting in
  partially observable stochastic domains,'' \emph{Artificial Intelligence},
  vol. 101, no.~1, pp. 99--134, 1998.

\bibitem{krishnamurthy_2016}
V.~Krishnamurthy, \emph{Partially Observed Markov Decision Processes: From
  Filtering to Controlled Sensing}.\hskip 1em plus 0.5em minus 0.4em\relax
  Cambridge University Press, 2016.

\bibitem{hazla2021bayesian}
J.~Hazla, A.~Jadbabaie, E.~Mossel, and M.~A. Rahimian, ``Bayesian decision
  making in groups is hard,'' \emph{Operations Research}, vol.~69, no.~2, pp.
  632--654, 2021.

\bibitem{omidshafiei2017deep}
S.~Omidshafiei, J.~Pazis, C.~Amato, J.~P. How, and J.~Vian, ``Deep
  decentralized multi-task multi-agent reinforcement learning under partial
  observability,'' in \emph{Proc. International Conference on Machine
  Learning}.\hskip 1em plus 0.5em minus 0.4em\relax PMLR, 2017, pp. 2681--2690.

\bibitem{gupta2017cooperative}
J.~K. Gupta, M.~Egorov, and M.~Kochenderfer, ``Cooperative multi-agent control
  using deep reinforcement learning,'' in \emph{Proc. AAMAS}, 2017, pp. 66--83.

\bibitem{foerster2018counterfactual}
J.~Foerster, G.~Farquhar, T.~Afouras, N.~Nardelli, and S.~Whiteson,
  ``Counterfactual multi-agent policy gradients,'' in \emph{Proc. {AAAI}
  Conference on Artificial Intelligence}, 2018, pp. 2974--2982.

\bibitem{moreno2018neural}
P.~Moreno, J.~Humplik, G.~Papamakarios, B.~A. Pires, L.~Buesing, N.~Heess, and
  T.~Weber, ``Neural belief states for partially observed domains,'' in
  \emph{NeurIPS Workshop on Reinforcement Learning under Partial
  Observability}, 2018, pp. 1--5.

\bibitem{gregor2018temporal}
K.~Gregor, G.~Papamakarios, F.~Besse, L.~Buesing, and T.~Weber, ``Temporal
  difference variational auto-encoder,'' in \emph{Proc. International
  Conference on Learning Representations}, 2019, pp. 1--17.

\bibitem{sayed_2022}
A.~H. Sayed, \emph{Inference and Learning from Data}.\hskip 1em plus 0.5em
  minus 0.4em\relax Cambridge University Press, 2022, 3 vols.

\bibitem{moreno2021neural}
P.~Moreno, E.~Hughes, K.~R. McKee, B.~A. Pires, and T.~Weber, ``Neural
  recursive belief states in multi-agent reinforcement learning,''
  \emph{arXiv:2102.02274}, 2021.

\bibitem{muglich2022}
D.~Muglich, L.~M. Zintgraf, C.~A.~S. De~Witt, S.~Whiteson, and J.~Foerster,
  ``Generalized beliefs for cooperative {AI},'' in \emph{Proc. International
  Conference on Machine Learning}, vol. 162, Jul 2022, pp. 16\,062--16\,082.

\bibitem{mao2020}
W.~Mao, K.~Zhang, E.~Miehling, and T.~Basar, ``Information state embedding in
  partially observable cooperative multi-agent reinforcement learning,'' in
  \emph{Proc. IEEE CDC}, 2020, pp. 6124--6131.

\bibitem{kar2013}
S.~Kar, J.~M.~F. Moura, and H.~V. Poor, ``${{\cal Q} {\cal D}}$-learning: A
  collaborative distributed strategy for multi-agent reinforcement learning
  through ${\rm consensus} + {\rm innovations}$,'' \emph{IEEE Transactions on
  Signal Processing}, vol.~61, no.~7, pp. 1848--1862, 2013.

\bibitem{zhang2018fully}
K.~Zhang, Z.~Yang, H.~Liu, T.~Zhang, and T.~Basar, ``Fully decentralized
  multi-agent reinforcement learning with networked agents,'' in \emph{Proc.
  International Conference on Machine Learning}.\hskip 1em plus 0.5em minus
  0.4em\relax PMLR, 2018, pp. 5872--5881.

\bibitem{cassano2021}
L.~Cassano, K.~Yuan, and A.~H. Sayed, ``Multiagent fully decentralized value
  function learning with linear convergence rates,'' \emph{IEEE Transactions on
  Automatic Control}, vol.~66, no.~4, pp. 1497--1512, 2021.

\bibitem{macua2021fully}
S.~V. Macua, I.~Davies, A.~Tukiainen, and E.~M. De~Cote, ``Fully distributed
  actor-critic architecture for multitask deep reinforcement learning,''
  \emph{The Knowledge Engineering Review}, vol.~36, pp. 1--30, 2021.

\bibitem{sha2022fully}
X.~Sha, J.~Zhang, K.~You, K.~Zhang, and T.~Ba{\c{s}}ar, ``Fully asynchronous
  policy evaluation in distributed reinforcement learning over networks,''
  \emph{Automatica}, vol. 136, p. 110092, 2022.

\bibitem{lin_stochastic_network_2021}
Y.~Lin, G.~Qu, L.~Huang, and A.~Wierman, ``Multi-agent reinforcement learning
  in stochastic networked systems,'' in \emph{Advances in Neural Information
  Processing Systems}, vol.~34, 2021, pp. 7825--7837.

\bibitem{wang2020}
G.~Wang, S.~Lu, G.~Giannakis, G.~Tesauro, and J.~Sun, ``Decentralized {TD}
  tracking with linear function approximation and its finite-time analysis,''
  in \emph{Advances in Neural Information Processing Systems}, 2020, pp.
  13\,762--13\,772.

\bibitem{sun2020finite}
J.~Sun, G.~Wang, G.~B. Giannakis, Q.~Yang, and Z.~Yang, ``Finite-time analysis
  of decentralized temporal-difference learning with linear function
  approximation,'' in \emph{Proc. International Conference on Artificial
  Intelligence and Statistics}, 2020, pp. 4485--4495.

\bibitem{lin2021}
Q.~Lin and Q.~Ling, ``Decentralized {TD(0)} with gradient tracking,''
  \emph{IEEE Signal Processing Letters}, vol.~28, pp. 723--727, 2021.

\bibitem{mahajan2016decentralized}
A.~Mahajan and M.~Mannan, ``Decentralized stochastic control,'' \emph{Annals of
  Operations Research}, vol. 241, no. 1-2, pp. 109--126, 2016.

\bibitem{malikopoulos2022team}
A.~A. Malikopoulos, ``On team decision problems with nonclassical information
  structures,'' \emph{IEEE Transactions on Automatic Control}, 2022.

\bibitem{yuksel2009stochastic}
S.~Yuksel, ``Stochastic nestedness and the belief sharing information
  pattern,'' \emph{IEEE Transactions on Automatic Control}, vol.~54, no.~12,
  pp. 2773--2786, 2009.

\bibitem{nayyar2013decentralized}
A.~Nayyar, A.~Mahajan, and D.~Teneketzis, ``Decentralized stochastic control
  with partial history sharing: A common information approach,'' \emph{IEEE
  Transactions on Automatic Control}, vol.~58, no.~7, pp. 1644--1658, 2013.

\bibitem{sutton2018reinforcement}
R.~S. Sutton and A.~G. Barto, \emph{Reinforcement Learning: An
  Introduction}.\hskip 1em plus 0.5em minus 0.4em\relax MIT Press, 2018.

\bibitem{tsitsiklis1996analysis}
J.~Tsitsiklis and B.~Van~Roy, ``Analysis of temporal-difference learning with
  function approximation,'' \emph{Advances in Neural Information Processing
  Systems}, vol.~9, 1996.

\bibitem{singh1994learning}
S.~P. Singh, T.~Jaakkola, and M.~I. Jordan, ``Learning without state-estimation
  in partially observable {Markovian} decision processes,'' in \emph{Proc.
  Machine Learning}, 1994, pp. 284--292.

\bibitem{rodriguez1999}
A.~Rodriguez, R.~Parr, and D.~Koller, ``Reinforcement learning using
  approximate belief states,'' in \emph{Advances in Neural Information
  Processing Systems}, vol.~12.\hskip 1em plus 0.5em minus 0.4em\relax MIT
  Press, 1999.

\bibitem{kimura1997reinforcement}
H.~Kimura, K.~Miyazaki, and S.~Kobayashi, ``Reinforcement learning in {POMDPs}
  with function approximation,'' in \emph{Proc. ICML}, vol.~97, 1997, pp.
  152--160.

\bibitem{cai2022reinforcement}
Q.~Cai, Z.~Yang, and Z.~Wang, ``Reinforcement learning from partial
  observation: Linear function approximation with provable sample efficiency,''
  in \emph{Proc. International Conference on Machine Learning}, 2022, pp.
  2485--2522.

\bibitem{li2021}
Y.~Li, Y.~Tang, R.~Zhang, and N.~Li, ``Distributed reinforcement learning for
  decentralized linear quadratic control: A derivative-free policy optimization
  approach,'' \emph{IEEE Transactions on Automatic Control}, pp. 6429--6444,
  2021.

\bibitem{wang2021distributed}
H.~Wang, S.~Lin, H.~Jafarkhani, and J.~Zhang, ``Distributed {Q}-learning with
  state tracking for multi-agent networked control,'' in \emph{Proc.
  International Conference on Autonomous Agents and Multi-Agent Systems}, 2021,
  pp. 1692--1694.

\bibitem{mahajan2012information}
A.~Mahajan, N.~C. Martins, M.~C. Rotkowitz, and S.~Y{\"u}ksel, ``Information
  structures in optimal decentralized control,'' in \emph{Proc. IEEE
  CDC}.\hskip 1em plus 0.5em minus 0.4em\relax IEEE, 2012, pp. 1291--1306.

\bibitem{saldi2022geometry}
N.~Saldi and S.~Y{\"u}ksel, ``Geometry of information structures, strategic
  measures and associated stochastic control topologies,'' \emph{Probability
  Surveys}, vol.~19, pp. 450--532, 2022.

\bibitem{arabneydi2015reinforcement}
J.~Arabneydi and A.~Mahajan, ``Reinforcement learning in decentralized
  stochastic control systems with partial history sharing,'' in \emph{Proc.
  American Control Conference (ACC)}, 2015, pp. 5449--5456.

\bibitem{nayyar2019common}
A.~Nayyar and D.~Teneketzis, ``Common knowledge and sequential team problems,''
  \emph{IEEE Transactions on Automatic Control}, vol.~64, no.~12, pp.
  5108--5115, 2019.

\bibitem{kayaalp2022dslw}
M.~Kayaalp, V.~Bordignon, S.~Vlaski, and A.~H. Sayed, ``Hidden {Markov}
  modeling over graphs,'' in \emph{Proc. IEEE Data Science and Learning
  Workshop (DSLW)}, 2022, pp. 1--6.

\bibitem{kayaalp_dist_bayesian}
M.~Kayaalp, V.~Bordignon, S.~Vlaski, V.~Matta, and A.~H. Sayed, ``Distributed
  {Bayesian} learning of dynamic states,'' \emph{arXiv:2212.02565}, 2022.

\bibitem{hlinka_2012}
O.~Hlinka, O.~Slučiak, F.~Hlawatsch, P.~M. Djurić, and M.~Rupp, ``Likelihood
  consensus and its application to distributed particle filtering,'' \emph{IEEE
  Transactions on Signal Processing}, vol.~60, no.~8, pp. 4334--4349, 2012.

\bibitem{battistelli_2014}
G.~Battistelli and L.~Chisci, ``{Kullback–Leibler} average, consensus on
  probability densities, and distributed state estimation with guaranteed
  stability,'' \emph{Automatica}, vol.~50, no.~3, pp. 707--718, 2014.

\bibitem{bandyopadhyay_2018}
S.~Bandyopadhyay and S.~Chung, ``Distributed {Bayesian} filtering using
  logarithmic opinion pool for dynamic sensor networks,'' \emph{Automatica},
  vol.~97, pp. 7--17, 2018.

\bibitem{sayed_2014}
A.~H. Sayed, ``{Adaptation, learning, and optimization over networks},''
  \emph{Foundations and Trends in Machine Learning}, vol.~7, no. 4-5, pp.
  311--801, July 2014.

\bibitem{nedic2009distributed}
A.~Nedic and A.~Ozdaglar, ``Distributed subgradient methods for multi-agent
  optimization,'' \emph{IEEE Transactions on Automatic Control}, vol.~54,
  no.~1, pp. 48--61, 2009.

\bibitem{dimakis2010gossip}
A.~G. Dimakis, S.~Kar, J.~M. Moura, M.~G. Rabbat, and A.~Scaglione, ``Gossip
  algorithms for distributed signal processing,'' \emph{Proc. IEEE}, vol.~98,
  no.~11, pp. 1847--1864, 2010.

\bibitem{di2016next}
P.~Di~Lorenzo and G.~Scutari, ``Next: In-network nonconvex optimization,''
  \emph{IEEE Transactions on Signal and Information Processing over Networks},
  vol.~2, no.~2, pp. 120--136, 2016.

\bibitem{chen2015learning}
J.~Chen and A.~H. Sayed, ``On the learning behavior of adaptive networks—part
  {I}: Transient analysis,'' \emph{IEEE Transactions on Information Theory},
  vol.~61, no.~6, pp. 3487--3517, 2015.

\bibitem{chen2015learning2}
------, ``On the learning behavior of adaptive networks—part {II}:
  Performance analysis,'' \emph{IEEE Transactions on Information Theory},
  vol.~61, no.~6, pp. 3518--3548, 2015.

\bibitem{kayaalp_dif_maml}
M.~Kayaalp, S.~Vlaski, and A.~H. Sayed, ``{Dif-MAML}: Decentralized multi-agent
  meta-learning,'' \emph{IEEE Open Journal of Signal Processing}, vol.~3, pp.
  71--93, 2022.

\bibitem{csiszar11}
I.~Csiszár and J.~Körner, \emph{Information Theory: Coding Theorems for
  Discrete Memoryless Systems}, 2nd~ed.\hskip 1em plus 0.5em minus 0.4em\relax
  Cambridge University Press, 2011.

\bibitem{sutton1988learning}
R.~S. Sutton, ``Learning to predict by the methods of temporal differences,''
  \emph{Machine Learning}, vol.~3, no.~1, pp. 9--44, 1988.

\bibitem{kolter2009regularization}
J.~Z. Kolter and A.~Y. Ng, ``Regularization and feature selection in
  least-squares temporal difference learning,'' in \emph{Proc. International
  Conference on Machine Learning}, 2009, pp. 521--528.

\bibitem{hoffman2011regularized}
M.~W. Hoffman, A.~Lazaric, M.~Ghavamzadeh, and R.~Munos, ``Regularized least
  squares temporal difference learning with nested $\ell_2$ and $\ell_1$
  penalization,'' in \emph{Proc. European Workshop on Reinforcement
  Learning}.\hskip 1em plus 0.5em minus 0.4em\relax Springer, 2011, pp.
  102--114.

\bibitem{farebrother2018generalization}
J.~Farebrother, M.~C. Machado, and M.~Bowling, ``Generalization and
  regularization in {DQN},'' \emph{arXiv:1810.00123}, 2018.

\bibitem{cobbe19quantifying}
K.~Cobbe, O.~Klimov, C.~Hesse, T.~Kim, and J.~Schulman, ``Quantifying
  generalization in reinforcement learning,'' in \emph{Proc. International
  Conference on Machine Learning}, vol.~97, 09--15 Jun 2019, pp. 1282--1289.

\bibitem{kiran2022}
B.~R. Kiran, I.~Sobh, V.~Talpaert, P.~Mannion, A.~A.~A. Sallab, S.~Yogamani,
  and P.~Pérez, ``Deep reinforcement learning for autonomous driving: A
  survey,'' \emph{IEEE Transactions on Intelligent Transportation Systems},
  vol.~23, no.~6, pp. 4909--4926, 2022.

\bibitem{biyik2022learning}
E.~B{\i}y{\i}k, D.~P. Losey, M.~Palan, N.~C. Landolfi, G.~Shevchuk, and
  D.~Sadigh, ``Learning reward functions from diverse sources of human
  feedback: Optimally integrating demonstrations and preferences,'' \emph{The
  International Journal of Robotics Research}, vol.~41, no.~1, pp. 45--67,
  2022.

\bibitem{li2019aaga}
T.~Li, H.~Fan, J.~García, and J.~M. Corchado, ``Second-order statistics
  analysis and comparison between arithmetic and geometric average fusion:
  Application to multi-sensor target tracking,'' \emph{Information Fusion},
  vol.~51, pp. 233--243, 2019.

\bibitem{kayaalp2022arithmetic}
M.~Kayaalp, Y.~Inan, E.~Telatar, and A.~H. Sayed, ``On the arithmetic and
  geometric fusion of beliefs for distributed inference,''
  \emph{arXiv:2204.13741}, April 2022.

\bibitem{bertsekas2015parallel}
D.~Bertsekas and J.~Tsitsiklis, \emph{Parallel and Distributed Computation:
  Numerical Methods}.\hskip 1em plus 0.5em minus 0.4em\relax Athena Scientific,
  2015.

\bibitem{acemoglu_2011}
D.~Acemoglu, M.~A. Dahleh, I.~Lobel, and A.~Ozdaglar, ``Bayesian learning in
  social networks,'' \emph{The Review of Economic Studies}, vol.~78, no.~4, pp.
  1201--1236, 2011.

\bibitem{garey1979computers}
M.~R. Garey and D.~S. Johnson, \emph{Computers and Intractability}.\hskip 1em
  plus 0.5em minus 0.4em\relax San Francisco: Freeman, 1979, vol. 174.

\bibitem{resnick2002}
S.~I. Resnick, \emph{Adventures in Stochastic Processes}.\hskip 1em plus 0.5em
  minus 0.4em\relax Birkhäuser, 2002.

\bibitem{jorge2016learning}
E.~Jorge, M.~K{\aa}geb{\"a}ck, F.~D. Johansson, and E.~Gustavsson, ``Learning
  to play guess who? and inventing a grounded language as a consequence,''
  \emph{arXiv:1611.03218}, 2016.

\bibitem{polyanskiy_2017}
Y.~Polyanskiy and Y.~Wu, ``Strong data-processing inequalities for channels and
  {Bayesian} networks,'' in \emph{Convexity and Concentration}.\hskip 1em plus
  0.5em minus 0.4em\relax New York, NY: Springer New York, 2017, pp. 211--249.

\bibitem{bretagnolle_huber}
J.~Bretagnolle and C.~Huber, ``Estimation des densit{\'e}s : Risque minimax,''
  in \emph{S{\'e}minaire de Probabilit{\'e}s XII}, C.~Dellacherie, P.~A. Meyer,
  and M.~Weil, Eds.\hskip 1em plus 0.5em minus 0.4em\relax Berlin, Heidelberg:
  Springer Berlin Heidelberg, 1978, pp. 342--363.

\bibitem{pillai2005}
S.~Pillai, T.~Suel, and S.~Cha, ``The {Perron-Frobenius} theorem: some of its
  applications,'' \emph{IEEE Signal Processing Magazine}, vol.~22, no.~2, pp.
  62--75, 2005.

\end{thebibliography}

\end{document}